\documentclass[11pt]{article}

\usepackage[preprint]{acl}

\usepackage{times}
\usepackage{latexsym}

\usepackage[T1]{fontenc}

\usepackage[utf8]{inputenc}

\usepackage{microtype}

\usepackage{inconsolata}

\usepackage{graphicx}

\usepackage{amsmath}
\usepackage{amsfonts}
\usepackage{tcolorbox}
\usepackage{listings}
\usepackage{graphicx}
\usepackage{booktabs}
\usepackage{multirow}
\usepackage{multicol}
\usepackage{subcaption}
\usepackage{adjustbox}
\usepackage[capitalise]{cleveref}
\usepackage[table]{xcolor}
\tcbuselibrary{listings,breakable}
\usepackage{float}

\newcommand{\interns}[0]{InternVL3.5~}
\newcommand{\intern}[0]{InternVL3.5}
\newcommand{\llavaos}[0]{LLaVA-OneVision~}
\newcommand{\llavao}[0]{LLaVA-OneVision}
\newcommand{\qwens}[0]{Qwen2.5-VL~}
\newcommand{\qwen}[0]{Qwen2.5-VL}
\newcommand{\molmos}[0]{Molmo~}
\newcommand{\molmo}[0]{Molmo}

\title{Getting to the Point:\\Pointing Improves LVLMs at Counting}

\author{
    Simone Alghisi, Massimo Rizzoli, Seyed Mahed Mousavi, Giuseppe Riccardi \\
    Signals and Interactive Systems Lab, University of Trento, Italy\\
    \texttt{ \{s.alghisi, giuseppe.riccardi\}@unitn.it}
}

\begin{document}
\maketitle
\begin{abstract}
    Pointing-based methods decompose complex tasks as sequential grounding and reasoning steps. Given a query, the model first grounds the relevant objects by generating their coordinates, and then predicts an answer conditioned on these points. While this approach has been shown to increase the performance of Large Vision-Language Models (LVLMs), it remains unclear why and how it improves the models' visual reasoning. In this work, we evaluate pointing-based methods in the task of zero-shot counting in visual scenes. We experiment with multiple fine-tuning and training-free approaches on state-of-the-art LVLMs, and compare them with \textit{Point-then-Count} (PtC), where models first generate point coordinates for the target objects and then predict their count. Our results show that PtC achieves the highest accuracy among the evaluated approaches, with predicted points correctly grounded in the image in more than 94\% of cases (based on F1-score). Mechanistic analyses show that gains arise from spatial information encoded in the predicted coordinates. Nevertheless, grounding performance varies across image regions, revealing spatial biases. Finally, the results indicate that PtC improves out-of-distribution generalization on both synthetic and real data, suggesting the potential of coordinates to help LVLMs improve their counting skills\footnote{We release all the material at \url{https://github.com/sislab-unitn/Getting-to-the-Point}}.
\end{abstract}

\section{Introduction}
\label{sec:intro}
Large Vision-Language Models (LVLMs) have achieved competitive performance on several multimodal tasks, such as visual question answering, image captioning, and optical character recognition~\cite{Deitke_2025_CVPR, bai2025qwen2, wang2025internvl3}.
However, prior work has shown that LVLMs struggle on complex cognitive tasks that require spatial~\cite{rizzoli-etal-2025-civet, kamath2023s}, temporal~\cite{Rahmanzadehgervi_2024_ACCV}, and counting~\cite{Fu_2025_CVPR, vo2025vision} capabilities.
Moreover, while some tasks require explicit visual evidence (e.g., grounded image captioning), in other settings LVLMs do not ground their answer in the image~\cite{10.1007/978-3-031-72775-7_2}, making it difficult for humans to interpret.

To improve the accuracy and explainability of LVLMs, pointing-based methods~\cite{Deitke_2025_CVPR} have recently proposed decomposing complex tasks into two explicit sequential steps. Given a natural-language query about the objects in an image, the model first grounds the objects mentioned in the query by predicting their 2D coordinates in the image. Then, it generates an answer to the query, conditioned on these grounded coordinates.
Recent studies have shown that pointing can improve final-task accuracy in counting~\cite{Deitke_2025_CVPR}, spatial reasoning~\cite{yang2025poivre}, robotic affordance prediction~\cite{yuan2025robopoint}, and document understanding~\cite{ni2025point}. 
However, several key questions remain unresolved: i) Does pointing encourage learning skills (e.g., counting) rather than narrow tasks (e.g., counting in a fixed range); ii) Are coordinates grounded in the visual input and therefore viable as explanations; and iii) What mechanism makes pointing improve performance?
Addressing these questions is crucial for developing models capable of learning skills and producing outputs that are reliable and interpretable. 

In this work, we study pointing in the context of zero-shot counting in visual scenes~\cite{dai2024referring, alghisi2025re, zhang2024good}, a cognitive task that serves as a benchmark for multi-step reasoning.
We decompose counting into a two-step process: 1) grounding all target instances in the image that satisfy a natural language query, and 2) aggregating this information to return the number of target objects. 
This makes counting an ideal testbed for studying pointing, as the task naturally separates target grounding from subsequent aggregation.
We evaluate four state-of-the-art LVLMs under multiple fine-tuning and training-free approaches, including direct counting~\cite{acharya2019tallyqa, Rahmanzadehgervi_2024_ACCV} and reasoning~\cite{jaech2024openai}, and compare them with \textit{Point-then-Count} (PtC)~\cite{Deitke_2025_CVPR}, where LVLMs generate the target objects' coordinates, followed by their count.
To understand which approach better supports skill learning, we evaluate LVLMs' Out-of-Distribution (OOD) generalization using images containing more targets than were observed during fine-tuning.
For PtC models, we also study whether objects' coordinates can serve as suitable visual explanations by assessing how grounded these coordinates are in the image and their contributions towards the final prediction.
Finally, to mitigate data contamination, reduce exploitable correlations during testing and perform fine-grained analyses, we evaluate LVLMs by generating synthetic datasets, balanced across object categories, labels, and spatial configurations. To ensure that our findings extend beyond synthetic settings, we also evaluate model performance on real-world data.

We frame our work around three research questions:
\begin{enumerate}
    \item \textbf{Does Pointing Improve Task Performance, Encouraging Skill Learning in Counting?}
    We assess four state-of-the-art LVLMs under multiple counting approaches and evaluate their performance across multiple settings.
    We find that point supervision improves performance in unseen scenarios with more target objects than those observed during fine-tuning on both synthetic and real-world data. Moreover, deriving the final count from the number of coordinates achieves the highest accuracy.
    
    \item \textbf{Can Coordinates Serve as Valid and Accurate Visual Explanations?}
    By comparing predicted coordinates with ground-truth annotations on synthetic data, we show that these points are grounded in more than 94\% of cases.
    However, grounding performance for some models is not uniform across the image, hinting at potential pre-training biases or limitations in the attention mechanism.
    
    \item \textbf{Why Pointing Improves Counting Performance?}
    When we replace each coordinate with an irrelevant token, no model achieves the same OOD generalization performance on synthetic data, underscoring the importance of spatial information for generalization. Further ablations reveal that PtC models mostly rely on coordinates to compute the total number of objects, disregarding the visual modality.
\end{enumerate}

Our findings suggest that incorporating pointing as an intermediate task changes what LVLMs learn: models trained to generate grounded coordinates appear to acquire a more general counting procedure and expose intermediate evidence that can be verified. At the same time, the remaining failure modes indicate substantial room for improvement, motivating training paradigms that explicitly decompose perception and reasoning to better support skill learning and more interpretable outputs.

\section{Literature Review}
\label{sec:literature}
\textbf{Decomposing complex cognitive tasks} into intermediate sub-tasks has been shown to benefit humans and animals~\cite{KRUEGER2009380}, as well as deep learning models~\cite{wies2023subtask,NEURIPS2022_9d560961, nye2022show, zhou2023leasttomost, sprague2025to}. In the vision-language domain, decomposition has been adopted to separate \emph{perception} from \emph{reasoning}~\cite{NEURIPS2018_5e388103, Andreas_2016_CVPR}. Pointing-based methods~\cite{Deitke_2025_CVPR} have been proposed as an intermediate subtask for LVLMs, and have proven effective in counting~\cite{Deitke_2025_CVPR, clark2026molmo2}, document understanding~\cite{ni2025point}, spatial reasoning~\cite{yang2025poivre} and visual question answering~\cite{Man_2025_CVPR}.
However, prior work indicates that LVLMs produce intermediate groundings that are poorly aligned and may contain factual errors~\cite{wu2025grounded, 10.1007/978-3-031-72775-7_2}, motivating evaluation protocols that focus on the intermediate evidence~\cite{Xia_2025_ICCV}.
Although datasets with intermediate grounding annotations exist~\cite{lei-etal-2020-tvqa, wu2025grounded}, they may exhibit positional biases~\cite{kirillov2023segment}, thus limiting their usefulness for evaluation.

\textbf{Counting} requires a model to output the total number of target objects in an image. 
While early work focused on counting fixed target objects~\cite{liang2022end, xie2018microscopy, lin2022boosting}, recent approaches generalize across classes by specifying the target in a few-shot~\cite{JEON2025111276, Ranjan_2021_CVPR, he2024few, you2023few} or zero-shot manner~\cite{Liu_2025_ICCV, Paiss_2023_ICCV, Jiang_2023_CLIP-Count, Dai_2024_CVPR, campbell2024understanding}. 
While LVLMs are often (pre-)trained on multiple counting benchmarks~\cite{Deitke_2025_CVPR, beyer2024paligemmaversatile3bvlm}, recent work shows that they can struggle even with small counts~\cite{alghisi2025re, Rahmanzadehgervi_2024_ACCV}, suggesting limited generalization.
PtC alleviates this by providing coordinates as intermediate supervision~\cite{Deitke_2025_CVPR}, yielding higher accuracy.

\section{Experimental Setting}
\label{sec:setting}
We investigate the role of pointing in the zero-shot counting task on four state-of-the-art LVLMs. Unlike prior studies that primarily emphasize accuracy~\cite{Deitke_2025_CVPR, clark2026molmo2, bai2025qwen2}, our work focuses on explaining why coordinates help and what is the mechanism behind these gains.
We evaluate each model on synthetic and real-world data across multiple counting approaches, comparing them with PtC in both in-distribution (ID) and out-of-distribution (OOD) settings to identify which method performs best and understand why.

\subsection{Task: Zero-Shot Counting}
\label{subsec:task}
Counting requires determining the number of instances of a target object in an image. 
In a zero-shot setting~\cite{beyer2024paligemmaversatile3bvlm, Deitke_2025_CVPR, zhang2024good, campbell2024understanding}, the target object is specified via a natural-language query.
Formally, given:
\begin{itemize}
    \item an image $\mathcal{I} \in \mathbb{R}^{H \times W \times C}$ with height $H$, width $W$, and $C$ channels depicting some objects;
    \item a query $q$ in natural language that specifies the \textbf{class} and (optionally) the \underline{attributes} of the target object~\cite{alghisi2025re} (e.g., {\em How many \underline{blue} \textbf{stars} are there?});
    \item a model $\mathcal{M}$, capable of processing an image $\mathcal{I}$ and a query $q$ in natural language.
\end{itemize}
we assess the counting capabilities of $\mathcal{M}$ by comparing the generated answer $\hat{y} = \mathcal{M}(\mathcal{I}, q)$ with $y$, where $y \in \mathbb{N}_0$ corresponds to the number of targets objects in $\mathcal{I}$ that match the class and attributes specified in $q$.
To ensure $\hat{y}$ follows our evaluation format, we employ a regex to remove any additional tokens (e.g., ``The answer is\ldots'', or ``Answer: \ldots'') and match the first valid answer. Further details are provided in \cref{subsec:answer-extraction}.

\subsection{Synthetic Data}
\label{subsec:synth-data}
Existing datasets are often heavily imbalanced, both in terms of classes~\cite{alghisi2025re} (e.g., ``person'' frequently dominating the distribution) and counts~\cite{al2024unibench} (small numbers occur more often than larger ones). Most exhibit strong positional bias, with targets frequently appearing in the center~\cite{kirillov2023segment}, thereby limiting our understanding of their grounding capabilities. Curated subsets~\cite{Deitke_2025_CVPR, beyer2024paligemmaversatile3bvlm} are often small and span only a narrow range, making them unsuitable to test OOD generalization. Finally, widely used real~\cite{acharya2019tallyqa} and synthetic~\cite{Johnson_2017_CVPR} datasets may have been included in the training data of some models, raising the risk of contamination.

We addressed these issues by generating tailored synthetic datasets with the CIVET framework~\cite{rizzoli-etal-2025-civet}, which provides fine-grained control over image content, allowing us to specify the class, attributes, and position of each object.
Leveraging such control, we construct multiple datasets, ensuring exhaustive coverage in terms of object types and position, and uniform label distribution.
For each dataset, we define a sample as a tuple $(\mathcal{I}, q, \mathcal{C}, y)$, where:
\begin{itemize}
    \item $\mathcal{I}$ is an image, which we subdivide into an $N \times M$ grid\footnote{For our experiments, we set $N = M = 9$.}, where objects can be placed within each cell;
    \item $q$ is a natural-language query specifying the class and attributes of the target object;
    \item $\mathcal{C} = \{(n, m) \mid 0 \leq n < N,\; 0 \leq m < M\}$ is the set of grid coordinates $(n,m)$ of all target objects in $\mathcal{I}$;
    \item $y$ is the number of target objects contained in $\mathcal{I}$ and corresponds to the cardinality of $\mathcal{C}$ (i.e., $y = |\mathcal{C}|$)
\end{itemize}
Having a grid-based representation gives us exact coordinates for each target object and control over the spatial layout.
This makes the dataset ideal for training PtC models and for evaluating grounding without annotation errors.

We start by constructing a balanced dataset to count from one to nine, as prior work has shown that LVLMs exhibit limited counting ability even for small numerosities~\cite{al2024unibench, alghisi2025re}, and existing curated benchmarks adopt a similar scope~\cite{beyer2024paligemmaversatile3bvlm, Deitke_2025_CVPR}. To reduce ambiguity, we consider images containing only target objects, as this ensures that model errors can be attributed to counting failures rather than confusion between targets and irrelevant objects. We report additional experiments with distractors in~\cref{subsec:distractors-appendix}.

For the training and validation splits, we use 10 target objects: six colored plusses and four white shapes. To obtain a uniform label distribution, we generate 81 unique samples (i.e., with a different spatial configuration) for each target object and count in the range $[1,9]$. 
We then construct an \texttt{ID} test set using the same data-generation procedure as the training set, while varying both spatial configurations and target objects.
Specifically, we consider 24 held-out target objects, obtained by recombining the 4 object classes and 6 colors observed during training into unseen color-shape combinations.

Finally, to understand which approach leads to better generalization in counting, we construct an \texttt{OOD} test set with images depicting from 10 to 18 objects. We generate this dataset using the same target objects as in the \texttt{ID} test set, allowing us to isolate generalization to higher counts from changes in object categories and visual attributes. Additional details on dataset construction and split sizes are provided in~\cref{subsec:synth-construction-appendix}.

\subsection{Real-World Data}
To assess whether our findings extend beyond synthetic data, we also train and evaluate our models on real-world images.
As previously discussed, most existing counting benchmarks provide human annotations only over a limited count range~\cite{Deitke_2025_CVPR, beyer2024paligemmaversatile3bvlm}, making them unsuitable for studying generalization to higher counts.
We therefore adapt the OCID~\cite{8793917} dataset for counting: OCID contains real-world scenes in which objects are placed incrementally, yielding images with 1 to 20 objects.
Each object is associated with a ground-truth segmentation mask, from which we derive pointing coordinates $\mathcal{C}$ using the mask centroid. This makes OCID a well-suited real-world counterpart to our synthetic dataset, with a similar count range (1-20 instead of 1-18) and instance-level coordinates for evaluating pointing-based counting. Using images with 1 to 10 objects, we construct the train, validation, and \texttt{ID} test splits, and reserve images with 11 to 20 objects for an \texttt{OOD} test split. Further details on dataset construction and splits are provided in \cref{subsec:real-construction-appendix}.

\begin{table*}[t!]
    \centering
    \setlength{\tabcolsep}{5pt}
    \small
        \begin{tabular}{lcc cc}
            \toprule
            \multirow{2}{*}{\textbf{Model}} &
            \multicolumn{2}{c}{\textit{Synthetic}} &
            \multicolumn{2}{c}{\textit{Real-World}} \\
            \cmidrule(lr){2-3}\cmidrule(lr){4-5}
            & \textbf{DC} & \textbf{PtC}
            & \textbf{DC} & \textbf{PtC} \\
            \midrule
            \textit{\qwens 3B}
            & 98.11 {\scriptsize (\textcolor[HTML]{228B22}{\(\uparrow\) 31.33})}
            & \textbf{99.94} {\scriptsize (\textcolor[HTML]{228B22}{\(\uparrow\) 33.15})}
            & 70.00 {\scriptsize (\textcolor[HTML]{228B22}{\(\uparrow\) 27.62})}
            & \textbf{81.90} {\scriptsize (\textcolor[HTML]{228B22}{\(\uparrow\) 39.52})} \\

            \textit{\qwens 7B}
            & 93.02 {\scriptsize (\textcolor[HTML]{228B22}{\(\uparrow\) 15.33})}
            & \textbf{99.88} {\scriptsize (\textcolor[HTML]{228B22}{\(\uparrow\) 22.19})}
            & 73.57 {\scriptsize (\textcolor[HTML]{228B22}{\(\uparrow\) 19.76})}
            & \textbf{76.19} {\scriptsize (\textcolor[HTML]{228B22}{\(\uparrow\) 22.38})} \\

            \textit{\llavaos}
            & \textbf{99.81} {\scriptsize (\textcolor[HTML]{228B22}{\(\uparrow\) 12.61})}
            & 99.76 {\scriptsize (\textcolor[HTML]{228B22}{\(\uparrow\) 12.56})}
            & 69.05 {\scriptsize (\textcolor[HTML]{228B22}{\(\uparrow\) 29.53})}
            & \textbf{77.14} {\scriptsize (\textcolor[HTML]{228B22}{\(\uparrow\) 37.62})} \\

            \textit{\interns}
            & 99.81 {\scriptsize (\textcolor[HTML]{228B22}{\(\uparrow\) 13.20})}
            & \textbf{100.00} {\scriptsize (\textcolor[HTML]{228B22}{\(\uparrow\) 13.39})}
            & 74.52 {\scriptsize (\textcolor[HTML]{228B22}{\(\uparrow\) 19.52})}
            & \textbf{80.24} {\scriptsize (\textcolor[HTML]{228B22}{\(\uparrow\) 25.24})} \\
            \bottomrule
        \end{tabular}
    \caption{Accuracy (\%) on Synthetic and Real-World in-distribution (\texttt{ID}) data after DC or PtC fine-tuning. Values in parentheses denote absolute improvements over the corresponding pre-trained model. \textbf{Bold} values indicate the best result on synthetic and real-world, respectively. Overall, models benefit more from point supervision, suggesting PtC as a more robust approach across object counts. \textit{Notation: \textbf{DC}: Direct Counting, \textbf{PtC}: Point-then-Count}}
    \label{tab:id-synthetic-real-dc-ptc}
\end{table*}

\subsection{Approach}
\label{subsec:approach}
We assess LVLM's capability to count under different approaches, and compare their performance on unseen scenarios to understand which approach generalizes best and why.
Following the notation introduced in \cref{subsec:task,,subsec:synth-data}, each model $\mathcal{M}$ takes as input an image $\mathcal{I}$ and a natural-language query $q$ specifying the target object. We consider the following approaches:
\begin{itemize}
    \item \textbf{Direct Counting (DC)}~\cite{acharya2019tallyqa}: $\mathcal{M}$ directly predicts the number of target objects $\hat{y}$ based on $(\mathcal{I}, q)$.

    \item \textbf{Point-then-Count (PtC)}~\cite{Deitke_2025_CVPR}: $\mathcal{M}$ first outputs the coordinates of the target objects and subsequently generates the predicted count $\hat{y}$.

    \item \textbf{Coordinates (\# Coord.)}: taking inspiration from grounding/detection-based counting~\cite{Dai_2024_CVPR, huang2024point}, we consider a variant of PtC in which the final count is deterministically computed as the number of predicted coordinates (Sec. \ref{subsec:answer-extraction}).

    \item \textbf{List-then-Count (LtC)}~\cite{hou2025assessing}: a variant of PtC in which $\mathcal{M}$ first enumerates the target objects, optionally grounding each instance with a short description or location cue, and then outputs the predicted count $\hat{y}$.

    \item \textbf{Reasoning}~\cite{jaech2024openai}: an extension of Chain-of-Thought~\cite{NEURIPS2022_9d560961} in which, without the need for examples, $\mathcal{M}$ generates an intermediate reasoning trace before producing the predicted count $\hat{y}$.
\end{itemize}

To better understand the contribution of pointing-based supervision in counting, we additionally fine-tune LVLMs to count under a DC and PtC approach.
This allows us to compare models that are trained on the same set of image-question pairs, while differing only in whether they receive coordinate-based supervision.
We ensure that performance differences can be attributed to the learning objective by using the same training protocol.
For both approaches, models are optimized with the standard next-token prediction objective~\cite{Deitke_2025_CVPR} using LoRA~\cite{hu2022lora}.
At inference time, we generate the answers with greedy decoding, using each LVLM's default stopping criterion.
Additional details about training, inference, and hardware can be found in~\cref{sec:experimental-details-appendix}.

\subsection{Models}
\label{subsec:models}
To understand whether our findings generalize across LVLMs with different architectural and image pre-processing choices, we evaluate representative state-of-the-art models: \qwens 7B~\cite{bai2025qwen2}, \llavao-1.5 8B~\cite{an2025llava}, and \interns 8B~\cite{wang2025internvl3}. In addition, we include \qwens 3B to study the effect of LLM scaling. We focus our analysis on open-source LVLMs, as closed-source models cannot be fine-tuned and may rely on external tools, making it difficult to assess their actual counting capabilities.
LVLMs combine a vision encoder with an LLM (i.e., a text decoder) to perform vision-language tasks, but differ in how visual features are integrated into the language model. Both \llavaos and \interns employ a learned projection layer to map the visual information extracted by the encoder into the LLM embedding space, while in \qwens the visual representation is passed directly to the decoder. 
Regarding image pre-processing, \interns applies dynamic tiling, splitting each image into non-overlapping $448\times448$ crops. In contrast, \llavaos and \qwens support variable-resolution inputs, eliminating the need for explicit resizing or cropping. Additional details about the checkpoint used for each model can be found in~\cref{subsec:model-appendix}.

\begin{table*}[t!]
    \centering
    \small
        \begin{tabular}{lccccc ccc}
            \toprule
            \multirow{2}{*}{\textbf{Model}} &
            \multicolumn{5}{c}{\textit{Training-Free}} &
            \multicolumn{3}{c}{\textit{Fine-Tuning}} \\
            \cmidrule(lr){2-6}\cmidrule(lr){7-9}
            & \textbf{DC}
            & \textbf{PtC}
            & \textbf{\# Coord.}
            & \textbf{LtC}
            & \textbf{Reasoning}
            & \textbf{DC}
            & \textbf{PtC}
            & \textbf{\# Coord.} \\
            \midrule
            \textit{\qwens 3B}
            & \underline{20.67} & 2.98 & 2.26 & 3.30 & --
            & 30.86 & 19.44 & \textbf{80.20} \\

            \textit{\qwens 7B}
            & \underline{23.41} & 14.04 & 12.96 & 6.38 & --
            & 32.00 & 46.19 & \textbf{94.96} \\

            \textit{\llavaos 8B}
            & \underline{19.19} & 9.31 & 7.51 & 2.78 & 14.76
            & 30.20 & 72.22 & \textbf{92.18} \\

            \textit{\interns 8B}
            & \underline{46.76} & 24.38 & 16.04 & 13.28 & 20.32
            & 45.52 & 97.33 & \textbf{97.94} \\

            \textit{Gemma 4 E4B}
            & 22.48 & 16.56 & 16.20 & \underline{22.89} & 21.34
            & -- & -- & -- \\
            \bottomrule
        \end{tabular}
    \caption{Accuracy (\%) on the synthetic out-of-distribution (\texttt{OOD}) setting across training-free and fine-tuning approaches. \textbf{Bold} values indicate the highest accuracy among fine-tuned approaches, while \underline{underlined} indicates the highest among training-free ones. Overall, computing the count from the coordinates predicted by fine-tuned models yields the highest accuracy. \textit{Notation: \textbf{DC}: Direct Counting, \textbf{PtC}: Point-then-Count, \textbf{LtC}: List-then-Count}}
    \label{tab:synthetic-ood}
    \small
\end{table*}

\section{Experiments}
\label{sec:experiments}
We evaluate four state-of-the-art LVLMs to count under different approaches. To evaluate which approach encourages skill acquisition, we compare their performance in an OOD scenario. We further examine whether predicted coordinates can serve as valid and accurate visual explanations, and conduct a mechanistic analysis to understand how pointing influences counting behavior. Because all datasets are balanced across object types and count labels, we report our results using accuracy.

\subsection{Does Pointing Always Improve Performance?}
Although pointing has been shown to improve LVLMs' counting accuracy~\cite{Deitke_2025_CVPR}, it is unknown whether point supervision always outperforms DC.
Investigating this question is crucial for practical applications, as the additional computation time and resources required to generate target coordinates may outweigh modest accuracy gains. This analysis also provides useful information to select the appropriate fine-tuning strategy under varying accuracy-efficiency trade-offs.
We thus study whether fine-tuning a model with point supervision leads to better performance in an \texttt{ID} setting, and compare its performance against DC.

The results in \cref{tab:id-synthetic-real-dc-ptc} show that models generally benefit more from point supervision than from DC fine-tuning. This trend is particularly noticeable in the real-world setting, where PtC improves accuracy by up to $11.9\%$ over DC. On synthetic data, the advantage of PtC is less pronounced, with \qwens 3B, \llavao, and \interns achieving comparable performance under the two fine-tuning strategies. However, additional per-count analysis in the appendix (\cref{fig:baseline-f1}) reveals that, after DC fine-tuning, the performance for both \qwens models degrades as the number of target objects increases, especially between 1 and 8 objects. This non-monotonic pattern is not observed for PtC, where performance remains more stable. In addition, models fine-tuned for PtC exhibit lower variability across object types and, in some cases, greater robustness to increasing numbers of distractors.
We provide further analyses of compositional generalization and distractor robustness in \cref{subsec:compositional-appendix,subsec:distractors-appendix}.

Overall, these findings indicate that point supervision improves in-distribution counting accuracy, especially on real-world images. Moreover, even when DC and PtC achieve similar aggregate performance on synthetic data, PtC yields more stable behavior as the number of objects increases, making it a more reliable fine-tuning strategy for counting.

\subsection{Which Approach Encourages Skill Learning?}
\label{subsec:skill-learning}
Understanding how models behave in unseen scenarios is particularly important for deployment, where the number of target objects may exceed what the model observed during training. In addition, investigating which approach leads to higher generalization is crucial for training, as collecting balanced real-world data for high object counts is almost unfeasible. 
To this end, we assess which approach encourages skill learning by measuring their performance in an \texttt{OOD} setting, using images with at most twice the targets seen during training.

We first evaluate the approaches on synthetic data. 
Since Reasoning is time- and compute-intensive, evaluation on the full dataset is unfeasible. We therefore compare all approaches on a stratified subset of the data (11\%), and report results on the full \texttt{OOD} set for the best-performing techniques in the appendix (Tab. \ref{tab:synthetic-ood-appendix}).
Results in~\cref{tab:synthetic-ood} show that fine-tuning surpasses training-free approaches. In general, models fine-tuned under PtC achieve higher accuracy than those trained on DC, suggesting that pointing may teach a more robust counting strategy, capable of generalizing to higher counts. Interestingly, deriving the final count from the number of coordinates (\# Coord.) increases accuracy substantially, with 7 and 8B models exceeding 90\%. \qwens 3B also exhibits a large improvement, but its performance remains below larger models, suggesting that capacity may be important.

Regarding training-free approaches, prompting models to point (PtC and \# Coord) or enumerate (LtC) objects generally underperform DC. 
This suggests that, without task-specific fine-tuning, most LVLMs struggle to use grounding or enumeration as intermediate steps for counting, consistent with previous findings in dialogue tasks~\cite{10.1007/978-3-031-72775-7_2}.
A similar trend emerges for Reasoning, where \interns achieves less than half of its DC accuracy, indicating that reasoning traces do not necessarily improve counting.
To assess whether this limitation is specific to older models, we additionally evaluate Gemma 4, released in April 2026. Even for this newer model, Reasoning achieves performance comparable to DC, but requires substantially more generated tokens. Unlike the other models, Gemma slightly benefits from LtC, suggesting that newer LVLMs may be capable of exploiting intermediate outputs, even without fine-tuning.

We next examine whether these findings transfer to real-world data, focusing on fine-tuned models, which achieved the highest accuracy on the synthetic benchmark. Results in~\cref{tab:real-ood-dc-ptc-coord} show that, except for \qwens 3B, fine-tuning improves over the corresponding training-free DC baseline (shown in gray). Consistent with the synthetic setting (Tab. \ref{tab:synthetic-ood}), deriving the final count from the coordinates yields the largest gains, with \qwens 3B improving by $+29.03$ accuracy points over its fine-tuned DC counterpart. The performance gap with the synthetic setting is likely due to the greater complexity of real-world images, where objects may be stacked or partially occluded, and to the lack of spatial balance in the data. Interestingly, despite using a similar image pre-processing to \qwens and achieving high accuracy in the synthetic setting, coordinates provide a marginal improvement for \llavao, suggesting that it may require different fine-tuning conditions.

\begin{table}[t!]
    \centering
    \setlength{\tabcolsep}{5pt}
    \small
        \begin{tabular}{lcccc}
            \toprule
            \multirow{2}{*}{\textbf{Model}} &
            \multicolumn{3}{c}{\textit{Fine-tuning}} &\\
            \cmidrule(lr){2-4}
            & \textbf{DC} & \textbf{PtC} & \textbf{\# Coord.} & \cellcolor[HTML]{E2E2E2}\textbf{DC} \\
            \midrule
            \textit{\qwens 3B}   
            & 2.15  & 0.98  & \textbf{31.18}
            & \cellcolor[HTML]{E2E2E2}2.35 \\

            \textit{\qwens 7B}   
            & 8.82  & 25.68 & \textbf{33.53}
            & \cellcolor[HTML]{E2E2E2}6.47 \\

            \textit{\llavaos} 
            & 12.55 & 15.69 & \textbf{16.27}
            & \cellcolor[HTML]{E2E2E2}1.76 \\

            \textit{\interns} 
            & 18.82 & 25.88 & \textbf{30.98}
            & \cellcolor[HTML]{E2E2E2}8.24 \\
            \bottomrule
        \end{tabular}
    \caption{Accuracy (\%) on the real-world out-of-distribution (\texttt{OOD}) setting for fine-tuning approaches as well as training-free direct counting  (\colorbox[HTML]{E2E2E2}{DC}). Similar to its synthetic counterpart, using the coordinates to count yields the highest accuracy.}
    \label{tab:real-ood-dc-ptc-coord}
\end{table}

These results suggest the potential of point supervision to generalization to higher object counts. Differing from prior work~\cite{Deitke_2025_CVPR}, we observe that using the predicted coordinates to count can be advantageous beyond the training distribution, as the model's final answer is often inconsistent with the number of predicted points.

\subsection{Are Coordinates Valid and Accurate Visual Explanations?}
\label{subsec:coordinates}
While our results show that coordinates can improve LVLM accuracy, especially in the \texttt{OOD} setting, their usefulness as visual explanations depends on whether they are grounded in the image and consistent with the final answer. 
We therefore evaluate their reliability using three metrics: F1-score, which captures both missed targets and hallucinated points; exact match (EM), which assesses whether the entire set of coordinates is correct; and consistency (Cons.), which measures whether the number of coordinates matches the final count.

We focus on synthetic data as it is balanced across spatial positions, enabling more fine-grained analyses. We report results on the synthetic \texttt{OOD} setting in~\cref{tab:grounding-quality-ood}, and discuss the remaining settings in~\Cref{subsec:grounding-appendix}.
Results show that consistency varies substantially across models, indicating that the number of predicted coordinates often does not match the model's final count. This explains why deriving the count directly from the predicted coordinates outperforms PtC in~\cref{tab:synthetic-ood}. EM further shows that high counting accuracy does not always imply fully correct grounding. For example, \qwens 7B reaches 94\% accuracy in~\cref{tab:synthetic-ood}, but only 88\% EM, meaning that in roughly 6\% of cases at least one predicted coordinate is incorrect. Nevertheless, F1-score remains high across all models, suggesting that grounding errors typically affect only a small subset of target objects.

\begin{table}[t!]
    \centering
    \small
    \setlength{\tabcolsep}{5pt}
        \begin{tabular}{lccc}
            \toprule
            \textbf{Model} & \textbf{F1} & \textbf{EM}  & \textbf{Cons.} \\
            \midrule
            \textit{\qwens 3B}          & 94.52 & 76.07 & 20.56 \\
            \textit{\qwens 7B}          & 98.59 & 88.05 & 48.79 \\
            \textit{\llavaos}        & 96.91 & 80.88 & 77.72 \\
            \textit{\interns}        & 99.89 & 98.06 & 99.27 \\
            \bottomrule
        \end{tabular}
    \caption{Coordinate reliability (\%) on the synthetic out-of-distribution (\texttt{OOD}) setting, reported using F1-score, exact match (EM), and consistency (Cons.) Consistency varies substantially across models, showing that, despite achieving high F1-scores, the number of predicted coordinates does not always match the model's final count.
    }
    \label{tab:grounding-quality-ood}
\end{table}

\begin{figure*}[ht]
    \centering
    \includegraphics[width=0.95\linewidth]{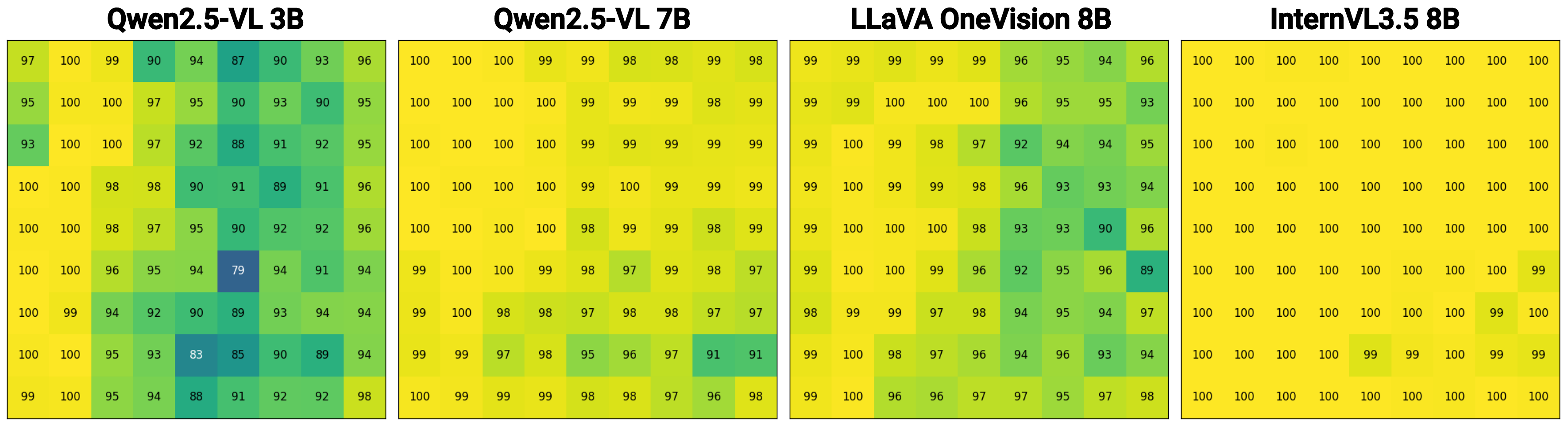}
    \caption{Cell-level F1-score (\%) for each model across our $9\times9$ grid on the synthetic out-of-distribution (\texttt{OOD}) setting. Except for \intern, the other models exhibit a higher F1-score for the left part of the image, showing the presence of spatial biases.}
    \label{fig:cell-level-f1-ood}
\end{figure*}

We then perform a fine-grained analysis by computing the F1-score for each cell in our $9\times9$ grid. \Cref{fig:cell-level-f1-ood} shows that both \qwens models and \llavaos exhibit spatial biases, with higher F1-score in the left portion of the image. Since \llavaos uses Qwen3 as its language backbone, this pattern may indicate a bias shared across this model family. In contrast, \interns maintains near-uniform and almost perfect F1-score across the image, which may be due to its pre-training or dynamic tiling strategy.

Overall, our findings show that predicted coordinates are grounded in the image in more than 94\% of cases, as measured by F1-score, supporting their use as visual explanations. However, the presence of spatial biases and reduced exact-match performance indicates that errors can concentrate in specific image regions, limiting the reliability of point-based explanations for some models.

\subsection{Why does Pointing Improve Performance?}
\label{subsec:why-ptc}
After showing that pointing improves LVLM counting accuracy and analyzing the reliability of the predicted coordinates, an important question remains: why does it help? 
Addressing this question is essential for understanding whether other tasks might also benefit from point supervision. 
To this end, we conduct a series of ablations to study why pointing improves LVLMs' counting accuracy, focusing on the synthetic \texttt{OOD} setting.
We hypothesize that coordinates provide explicit spatial cues that help the model disambiguate among object instances.
To test this, we fine-tune each model for PtC, but replace each coordinate with the token "X", and report results in~\cref{tab:ablations-a}. Our results show that removing the coordinates consistently reduces accuracy, indicating that spatial information helps LVLMs' generalize to higher object counts. The largest drop occurs for \qwens 7B, which, despite achieving 97\% on the \texttt{ID} setting (\cref{subsec:xft-appendix}), tends to repeatedly generate "X" tokens instead of meaningful outputs.

\begin{table}[!t]
    \centering
    \small
    \setlength{\tabcolsep}{3pt}
    \begin{subtable}[t]{0.60\linewidth}
        \centering
        \begin{tabular}{lc}
            \toprule
            \multirow{2}{*}{\textbf{Model}} & 
            \multirow{2}{*}{\textbf{X-FT}} \\
            \arrayrulecolor{white}
            \cmidrule(lr){2-2}
            \arrayrulecolor{black}
            & \\
            \midrule
            \textit{\qwens 3B}
            & \textcolor[HTML]{B22222}{\(\downarrow\) 60.50} \\
            
            \textit{\qwens 7B}
            & \textcolor[HTML]{B22222}{\(\downarrow\) 91.42} \\
            
            \textit{\llavaos 8B}
            & \textcolor[HTML]{B22222}{\(\downarrow\) 75.78} \\
            
            \textit{\interns 8B}
            & \textcolor[HTML]{B22222}{\(\downarrow\) 61.00} \\
            \bottomrule
        \end{tabular}
        \caption{}
        \label{tab:ablations-a}
    \end{subtable}
    \begin{subtable}[t]{0.38\linewidth}
        \centering
        \begin{tabular}{cc}
            \toprule
            \multicolumn{2}{c}{\textbf{Ablation}} \\
            \cmidrule(lr){1-2}
            \textbf{Image} & \textbf{Coord.}\\
            \midrule
            \textcolor[HTML]{B22222}{\(\downarrow\) 1.22} &
            \textcolor[HTML]{B22222}{\(\downarrow\) 92.13} \\
            
            \textcolor[HTML]{228B22}{\(\uparrow\) 0.12} &
            \textcolor[HTML]{B22222}{\(\downarrow\) 85.64} \\
            
            \textcolor[HTML]{B22222}{\(\downarrow\) 0.99} &
            \textcolor[HTML]{B22222}{\(\downarrow\) 93.65} \\
            
            \(=\) 0.00 &
            \textcolor[HTML]{B22222}{\(\downarrow\) 81.81} \\
            \bottomrule
        \end{tabular}
        \caption{}
        \label{tab:ablations-b}
    \end{subtable}
    \caption{Absolute change in accuracy under different ablations w.r.t. the PtC baseline. \textbf{(a)} reports the change on the synthetic out-of-distribution (\texttt{OOD}) setting (\cref{tab:synthetic-ood}) after X-FT, where each coordinate is replaced with an ``X'' during PtC training. \textbf{(b)} reports the change on the synthetic in-distribution (\texttt{ID}) setting (\cref{tab:id-synthetic-real-dc-ptc}) when ablating the input image or coordinates. Results highlight the importance of spatial information for generalization and the dominant role of textual coordinates.}
\end{table}

To better understand how models leverage coordinates when counting, we conduct two additional ablations. First, we assess the model's reliance on coordinates by replacing $\mathcal{I}$ with a black image while providing ground-truth coordinates $\mathcal{C}$ as input. Second, we perform a leave-one-out ablation in which we remove one coordinate at a time and measure whether the model can recover the missing point. We also perform additional activation patching experiments to show each layer contribution, and report the results in \cref{subsec:mechanistic-appendix}.
We perform our ablations on the \texttt{ID} setting, where models achieve the highest accuracy (Tab. \ref{tab:id-synthetic-real-dc-ptc}) and consistency (Tab. \ref{tab:grounding-quality}), making it well-suited to study the contribution of coordinates on the model's final count.
Results in~\cref{tab:ablations-b} reveal that removing the image has little effect ($<2\%$ drop) on performance, indicating that models generate the final count mostly based on the coordinates, disregarding the visual modality. Ablating the coordinates leads to a significant drop in accuracy ($>81\%$), indicating that models cannot recover when coordinate information is removed. This behavior resembles recent findings on reasoning models, where intermediate ``thinking'' tokens strongly influence the final prediction~\cite{xu2026more}. This suggests that, in PtC, coordinates may play a similar role.

Our mechanistic analysis indicates that spatial information helps LVLMs to generalize to higher object counts. Moreover, we find that models rely primarily on the textual coordinates to produce the final count, validating the idea that reasoning over a structured representation may be easier than using raw visual features~\cite{NEURIPS2018_5e388103,Man_2025_CVPR}.

\section{Conclusion}
We investigated the role of pointing-based methods in the zero-shot counting task by fine-tuning state-of-the-art LVLMs and comparing their performance with training-free methods. Our results show that Point-then-Count (PtC) achieves higher OOD generalization, suggesting that coordinate-based supervision can help LVLMs improve their counting skills. The generated coordinates are grounded in over 94\% of cases, but fine-grained analyses reveal spatial biases. Ablations show that the spatial information in the coordinates helps generalize to higher object counts. Moreover, we observe that PtC models primarily rely on the textual coordinates to count, disregarding the image content.
These findings motivate future training paradigms that more explicitly integrate grounding in reasoning, encouraging models to use visual evidence not only as an intermediate output but also as a reliable basis for subsequent reasoning.
In the future, we plan to extend our results to other tasks and explore novel training strategies that better integrate grounding with visual reasoning.

\section{Limitations}
We focus our analysis on open-source LVLMs up to 8B parameters, since proprietary models cannot be fine-tuned, and larger models exceed our computational resources.
Our study is also limited to the zero-shot counting task: understanding whether the findings hold in other vision-language tasks should be addressed in future work. In addition, although we validate our findings on real-world images, OCID serves as a relatively controlled scenario. Further evaluation is needed to assess whether these methods scale to more complex settings, including scenes with more variability, images containing a higher number of objects, and more diverse referring expressions.
Finally, our OOD evaluation focuses on one specific aspect of counting skill: generalization to higher object counts than those observed during training. This allows us to assess whether coordinate-based supervision improves count-range extrapolation, while leaving broader forms of counting generalization to future work.

\bibliography{custom}

\clearpage
\appendix

\section{Experimental Details}
\label{sec:experimental-details-appendix}
We complement the main paper by providing additional implementation details.

\subsection{Models}
\label{subsec:model-appendix}
We evaluate the following open-source LVLMs available on HuggingFace:
\begin{itemize}
    \item \href{https://huggingface.co/Qwen/Qwen2.5-VL-3B-Instruct}{\qwens 3B}
    \item \href{https://huggingface.co/Qwen/Qwen2.5-VL-7B-Instruct}{\qwens 7B}
    \item \href{https://huggingface.co/lmms-lab/LLaVA-OneVision-1.5-8B-Instruct}{\llavao-1.5 8B}
    \item \href{https://huggingface.co/OpenGVLab/InternVL3_5-8B}{\interns 8B}
    
\end{itemize}
These models were selected because they represent state-of-the-art open-source LVLMs with different architectural choices and image preprocessing strategies, as discussed in~\cref{subsec:models} of the main paper. In addition, we conduct preliminary experiments with \href{https://huggingface.co/allenai/Molmo-7B-O-0924}{Molmo-7B-O} and training-free experiments on \href{https://huggingface.co/google/gemma-4-E4B-it}{Gemma 4 E4B}.

\subsection{Training}
We fine-tune each LVLM to count under two approaches, DC and PtC. In \textbf{DC}, we optimize $\mathcal{M}$ to map the input pair $(\mathcal{I}, q)$ directly to the target count $y$. In \textbf{PtC}, we fine-tune $\mathcal{M}$ to first predict the coordinates $\mathcal{C}$ of the target objects based on $(\mathcal{I}, q)$, and subsequently generate the target count $y$. Specifically, we prepend $\mathcal{C}$ to the target count $y$, and optimize the model's parameters such that $\mathcal{M}(\mathcal{I}, q) =$ \textit{``Coordinates: }$(n_1, m_1), (n_2, m_2), \ldots, (n_y, m_y)$. \textit{Answer: }$y$\textit{''}. We sort the coordinates in a fixed left-to-right, top-to-bottom order, which was shown to outperform a random ordering~\cite{Deitke_2025_CVPR}.

We train each model using LoRA~\cite{hu2022lora} (rank $r=32$, scaling $\alpha=64$), inserting adapters into the query, key, and value matrices of LLM, vision encoder, and modality projection layer (when present). Models are loaded in FP16 precision and optimized using AdamW with default parameters and a learning rate of $1\times10^{-5}$ (which achieved the lowest validation perplexity in preliminary experiments). Training runs for up to 10 epochs, with early stopping (patience $=2$) based on validation perplexity. All experiments are conducted on a single NVIDIA A100 (80\,GiB) GPU with batch size 1, and images are processed using each model's default pre-processing pipeline. Under this configuration, each experiment requires at most two days.

\subsection{Inference}
\label{subsec:appendix-inference}
During inference, answers are generated using greedy decoding with each LVLM's default stopping criterion. The only exception is Reasoning, for which we use each model's recommended generation hyperparameters, such as top-$p$, top-$k$, and temperature, as specified in the corresponding Hugging Face documentation.

Regarding the token budget, for fine-tuned models, we set $\texttt{max\_new\_tokens}=5$ for DC and $\texttt{max\_new\_tokens}=1{,}000$ for PtC, ensuring that each model can generate both coordinates and the final answer regardless of differences across tokenizers. For training-free approaches, we set $\texttt{max\_new\_tokens}=3{,}000$ for PtC and LtC, and $\texttt{max\_new\_tokens}=32{,}768$ for Reasoning, to accommodate the longer reasoning trace.

We report the prompts used for each training-free approach in~\Cref{fig:dc-prompt,,fig:ptc-prompt-molmo,,fig:ptc-prompt-others,,fig:ltc-prompt,,fig:reasoning-prompt}. For Molmo, we use a prompt for PtC closely aligned with the one reported in its corresponding paper~\cite{Deitke_2025_CVPR}, to ensure that the model outputs the objects' coordinates. For the other models, we evaluate three different prompts on a subset of the validation set and retain the one that achieves the best results

\subsection{Answer and Coordinate Extraction}
\label{subsec:answer-extraction}
To extract the predicted count $\hat{y}$, we first convert textual numbers into digits (e.g., ``one'' $\rightarrow$ 1, ``five'' $\rightarrow$ 5, ``twenty'' $\rightarrow$ 20) and then apply a regular expression that extracts the first consecutive sequence of digits from the generated output. If no valid number can be matched, we return $-1$. For PtC models, the same procedure is applied after locating the \textit{Answer:} field. Predicted grounding coordinates are extracted using a separate regular expression that matches tuples of the form $(n,m)$ in the generated output, while discarding malformed pairs. The predicted coordinate set $\hat{\mathcal{C}}$ is evaluated against the ground-truth set $\mathcal{C}$ to assess grounding quality (\cref{subsec:coordinates}). In the Coordinates approach (\cref{subsec:approach}), we instead compute the predicted count directly as the number of extracted coordinates (i.e., $\hat{y} = |\hat{\mathcal{C}}|$).

\begin{figure*}[!ht]
    \centering
    \begin{tcolorbox}[
    title={Prompt for DC},
    colback=blue!5,
    colframe=blue!60,
    fonttitle=\bfseries
]
\small\ttfamily
Answer using as few words as possible.
\end{tcolorbox}
    \caption{Prompt used for training-free Direct Counting}
    \label{fig:dc-prompt}
\end{figure*}

\begin{figure*}[!ht]
    \centering
    \begin{tcolorbox}[
    title={Prompt for PtC (others)},
    colback=blue!5,
    colframe=blue!60,
    fonttitle=\bfseries
]
\small\ttfamily
Count the object(s) in the image. First generate the object's location(s)
(returning a pair of (x, y) coordinates between 0 and 100), then return the
total number of objects. Only answer with ``Coordinates: ... Answer:''.
\end{tcolorbox}
    \caption{Prompt used for training-free Point-then-Count for \llavao, \intern, and \qwen}
    \label{fig:ptc-prompt-others}
\end{figure*}

\begin{figure*}[!ht]
    \centering
    \begin{tcolorbox}[
        title={Prompt for PtC (\molmo)},
        colback=blue!5,
        colframe=blue!60,
        fonttitle=\bfseries
    ]
\small
\begin{verbatim}
Count by pointing.
\end{verbatim}
    \end{tcolorbox}
    \caption{Prompt template used for training-free Point-then-Count for \molmo.}
    \label{fig:ptc-prompt-molmo}
\end{figure*}

\begin{figure*}[!t]
    \centering
    \begin{tcolorbox}[
    title={Prompt for LtC},
    colback=blue!5,
    colframe=blue!60,
    fonttitle=\bfseries
]
\small\ttfamily
You are given an image. 

1. First, list all instances of the object mentioned in the question as a numbered list. Each list item must contain a brief identifier (e.g., position or distinguishing detail). 

2. Then, count the number of listed items. 

3. Finally, provide the final answer within <answer></answer>. The answer must match the number of listed items. \\

Output format:

<list> 

1. 

2. 

</list> 

<answer>N</answer>\\

Question:
\end{tcolorbox}
    \caption{Prompt used for List-then-Count}
    \label{fig:ltc-prompt}
\end{figure*}

\begin{figure*}[!t]
    \centering
    \begin{tcolorbox}[
    title={Prompt for Reasoning},
    colback=blue!5,
    colframe=blue!60,
    fonttitle=\bfseries
]
\small\ttfamily
Only answer with the numerical value.
\end{tcolorbox}
    \caption{Prompt used for Reasoning models.}
    \label{fig:reasoning-prompt}
\end{figure*}

\section{Dataset Construction}
We complement the main paper by providing additional information about synthetic and real-world dataset construction.

\subsection{Synthetic Datasets}
\label{subsec:synth-construction-appendix}
As image size can affect model accuracy, we generate each image $\mathcal{I}$ at a resolution of $672\times672$ pixels, following prior findings~\cite{rizzoli-etal-2025-civet}. For models that do not support this resolution, we apply their default image pre-processing strategy.

As discussed in~\cref{subsec:synth-data}, we subdivide each image into a $9\times9$ grid, where each cell corresponds to a $74\times74$ pixel region. Since $74\times9=666$, we add 3 pixels of padding on each side so that the grid is centered within the $672\times672$ image. Each object occupies an area of $64\times64$ pixels and can be placed in the center of a cell. To avoid occlusions, we place each object in a distinct cell. Within the grid, coordinates are represented as tuples $(n,m)$, where $n$ and $m$ denote the row and column indices, respectively. The top-left cell corresponds to $(0,0)$, while the bottom-right cell corresponds to $(8,8)$.

Following prior work~\cite{alghisi2025re}, we mitigate the effect of external confounding factors by considering images with a uniform black background, reducing the chance that irrelevant elements are mistaken for targets, and by placing objects in distinct cells to avoid occlusions. To ensure that the proposed setting does not induce a severe distribution shift, we run a preliminary experiment to assess the performance of LVLMs on the generated images and discuss the results in \cref{subsec:preliminary-appendix}. 

We generate images incrementally to minimize differences between images containing $i$ and $i-1$ target objects. Specifically, each image $\mathcal{I}_i$ is obtained by adding one target object to $\mathcal{I}_{i-1}$ (leaving the other objects unchanged). This ensures that consecutive images differ by exactly one object, allowing us to study how increasing the number of objects affects LVLMs while minimizing variation due to object position. We use the same procedure when introducing distractors into the scene.

Queries $q$ are generated using the template \textit{``How many \texttt{<color>} \texttt{<shape>}s are there?''}, where the placeholders are replaced with the color and shape of the target object (e.g., ``How many blue stars are there?'').

\begin{figure}[t]
    \centering
    \includegraphics[width=0.8\linewidth]{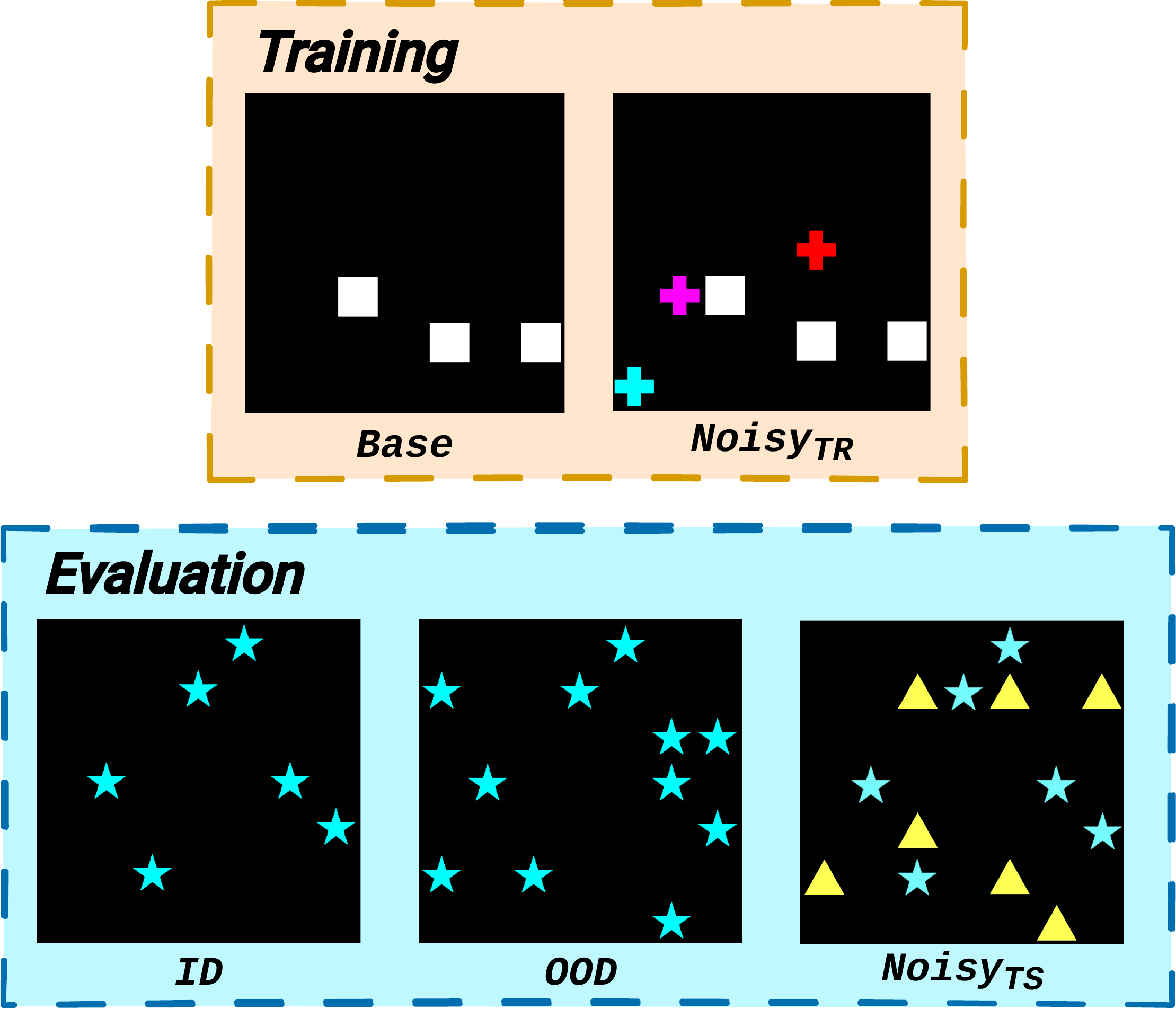}
    \caption{Visual overview of our synthetic datasets. In the main paper, we consider \texttt{Base} for training and \texttt{ID} and \texttt{OOD} for evaluation. \cref{subsec:distractors-appendix} discuss the results for \texttt{Noisy\textsubscript{TR}} and \texttt{Noisy\textsubscript{TS}}}
    \label{fig:synth-data}
\end{figure}

We describe our training and evaluation datasets below and provide a visual overview in \cref{fig:synth-data}.

\paragraph{Training data.}
As described in~\cref{subsec:synth-data}, we fine-tune models on images containing 1 to 9 target objects. To avoid overfitting on a specific class-attribute combination, we consider 10 different target objects: six colored plusses (i.e., \textit{red}, \textit{green}, \textit{blue}, \textit{cyan}, \textit{magenta}, \textit{yellow}) and four white shapes (i.e., \textit{circle}, \textit{square}, \textit{star}, \textit{triangle}).

The \texttt{Base} training setting, described in~\cref{subsec:synth-data}, contains 7,290 samples, obtained from 9 count labels, 10 target objects, and 81 spatial configurations. We split this dataset into 4,860 training samples and 2,430 validation samples.

We also construct a noisier training setting, \texttt{Noisy\textsubscript{TR}}, to assess whether exposure to irrelevant objects during fine-tuning improves robustness. Starting from each image $\mathcal{I}$ in the \texttt{Base} setting, we add $d \in \{1,2,3\}$ distractors to randomly selected empty cells. Each distractor is sampled uniformly from the set of non-target objects. To avoid introducing reasoning shortcuts, we enforce a uniform distribution over both the number and type of distractors. The resulting dataset has the same size as \texttt{Base}, allowing us to isolate the effect of increased scene complexity on LVLM learning without confounding it with additional training data.

\paragraph{Evaluation data.}
The \texttt{ID} and \texttt{OOD} test sets follow the construction described in~\cref{subsec:synth-data}. Both contain 17,496 samples and use the same 24 held-out target objects, obtained from all combinations of 4 shapes and 6 colors. The \texttt{ID} set contains images with 1 to 9 target objects, while the \texttt{OOD} set contains images with 10 to 18 target objects.

We further construct a distractor-based test set, \texttt{Noisy\textsubscript{TS}}, to evaluate robustness to irrelevant objects. To isolate the effect of distractors, we start from the \texttt{ID} images and add distractors while keeping the target count and target positions fixed. Since exhaustively evaluating all target-distractor pairs is computationally infeasible, we fix the target object to \textit{blue star}, which achieves the highest counting accuracy in the \texttt{ID} setting (\cref{subsec:compositional-appendix}). We use the remaining 23 class-attribute combinations as distractors, allowing us to marginalize over distractor types. To study the effect of distractor quantity, we divide the dataset into nine segments, where segment $d \in \{1,\ldots,9\}$ contains images with exactly $d$ distractors. For each image, we instantiate three distractor spatial configurations by placing the distractors in different empty cells. Each segment contains 50,301 samples, for a total of 452,709 samples.

\subsection{Real-World Dataset}
\label{subsec:real-construction-appendix}
To evaluate whether our findings extend beyond synthetic images, we construct a real-world counting benchmark from OCID~\cite{8793917}. OCID provides RGB images, instance-level segmentation masks, and scenes (i.e., sequences of images in which objects are incrementally added), allowing us to obtain images with different object counts under similar scene conditions.

We split OCID at the scene level to avoid overlap between train, validation, and test images. The split is stratified across floor and table scenes, and we include both top and bottom camera views when available. Images containing 1 to 10 objects are used for the train, validation, and \texttt{ID} test splits, yielding 1160, 200, and 420 samples, respectively. Images containing 11 to 20 objects are reserved for the \texttt{OOD} split, resulting in 510 samples.

For each image $\mathcal{I}$, we derive ground-truth coordinates $\mathcal{C}$ from the instance-level segmentation masks. For each object, we first compute the mask centroid and then select the valid mask pixel closest to it as the representative point. This ensures that the coordinate lies on the target object itself, avoiding cases where the centroid falls outside the visible mask or on an occluding object. Following prior work~\cite{Deitke_2025_CVPR}, we normalize coordinates to the $[0,100]$ range and keep one decimal digit. Since segmentation masks do not provide object-class labels, we use \textit{``How many objects are there?''} as the query $q$.

Due to the limited size of the training split, we apply data augmentation to the training images (i.e., color jittering, random crops, translations, rotations, and resizing) while updating the corresponding normalized coordinates. This process yields a total of 12,760 training samples.

\begin{table}[t]
    \centering
    \setlength{\tabcolsep}{10pt}
    \small
    \begin{tabular}{lcc}
        \toprule
        \textbf{Model} & \textbf{DC} & \textbf{PtC} \\
        \midrule
        \textit{\qwens 3B}   & \textbf{66.79} & 45.71 \\
        \textit{\qwens 7B}   & 77.69 & \textbf{78.42} \\
        \textit{\llavaos 8B} & \textbf{87.20} & 62.91 \\
        \textit{\interns 8B} & \textbf{86.61} & 77.42 \\
        \textit{\molmos 7B}  & 53.06 & \textbf{99.95} \\
        \bottomrule
    \end{tabular}
        \caption{Accuracy (\%) of each on the synthetic \texttt{ID} setting for training-free DC and PtC. Values in \textbf{bold} indicate the higher result between the two approaches. Overall, all models achieve more than 66\% accuracy, suggesting that our images are compatible with their (pre-)training distribution. On PtC, only \molmos achieves near-perfect accuracy, while other models either degrade or show little change.}
    \label{tab:baseline-pre-trained}
\end{table}

\begin{table}[t]
    \centering
    \small
    \begin{adjustbox}{width=\columnwidth}
        \begin{tabular}{lcccc}
            \toprule
            & \textit{Circles} & \textit{Squares} & \textit{Triangles} & \textit{Stars} \\
            \midrule
            \textit{R}
            & 99.86 {\scriptsize $\pm$ 0.11}
            & 99.83 {\scriptsize $\pm$ 0.34}
            & 99.97 {\scriptsize $\pm$ 0.07}
            & 99.93 {\scriptsize $\pm$ 0.08} \\
            
            \textit{G}
            & 99.90 {\scriptsize $\pm$ 0.07}
            & 99.83 {\scriptsize $\pm$ 0.21}
            & 99.93 {\scriptsize $\pm$ 0.14}
            & \textbf{100 {\scriptsize $\pm$ 0.00}} \\
            
            \textit{B}
            & 99.86 {\scriptsize $\pm$ 0.11}
            & 99.59 {\scriptsize $\pm$ 0.40}
            & 99.93 {\scriptsize $\pm$ 0.08}
            & 99.93 {\scriptsize $\pm$ 0.08} \\
            
            \textit{C}
            & 99.90 {\scriptsize $\pm$ 0.07}
            & 99.86 {\scriptsize $\pm$ 0.19}
            & 99.97 {\scriptsize $\pm$ 0.07}
            & 99.97 {\scriptsize $\pm$ 0.07} \\
            
            \textit{M}
            & 99.97 {\scriptsize $\pm$ 0.07}
            & 99.66 {\scriptsize $\pm$ 0.29}
            & 99.97 {\scriptsize $\pm$ 0.07}
            & 99.97 {\scriptsize $\pm$ 0.07} \\
            
            \textit{Y}
            & 99.90 {\scriptsize $\pm$ 0.07}
            & 99.79 {\scriptsize $\pm$ 0.26}
            & \textbf{100 {\scriptsize $\pm$ 0.00}}
            & 99.97 {\scriptsize $\pm$ 0.07} \\
            \bottomrule
        \end{tabular}
    \end{adjustbox}
    \caption{Accuracy (\%) across object shape and color combinations on the synthetic \texttt{ID} setting for fine-tuned PtC Models. Values report mean accuracy with standard deviation across models. Results in \textbf{bold} indicate the color-shape combination(s) with the highest accuracy and lower standard deviation.  \textit{Notation: \textbf{R}: red, \textbf{G}: green, \textbf{B}: blue, \textbf{C}: cyan, \textbf{M}: magenta, \textbf{Y}: yellow.}}
    \label{tab:shape_color_compositionality-PtC}
\end{table}

\section{Preliminary Results}
\label{subsec:preliminary-appendix}
To ensure that the setting discussed in~\cref{subsec:synth-data,,subsec:synth-construction-appendix} does not induce a severe distribution shift, we run a preliminary experiment to assess the performance of LVLMs on the generated images. The results in~\cref{tab:baseline-pre-trained} show that all pre-trained models achieve more than 66\% accuracy (best results between DC and PtC) on the \texttt{ID} setting, suggesting that our images are compatible with their pre-training distribution. When asked to first point to the objects, \molmo's performance almost doubles, reaching near-perfect accuracy and indicating that the model benefits considerably from point supervision. In contrast, the performance of \qwens 3B, \llavao, and \interns decreases, suggesting that these models may not have been pre-trained to ground objects or to use such information for counting. The only exception is \qwens 7B, which achieves only a negligible gain in performance ($<2\%$). While this may indicate that the model was trained to use grounding as an intermediate subtask (as mentioned in its report~\cite{bai2025qwen2}), the predicted coordinates do not follow the prompting schema and are, in most cases, out of bounds, making them not viable as visual explanations.

\begin{figure*}[!h]
    \centering
    \includegraphics[width=0.8\linewidth]{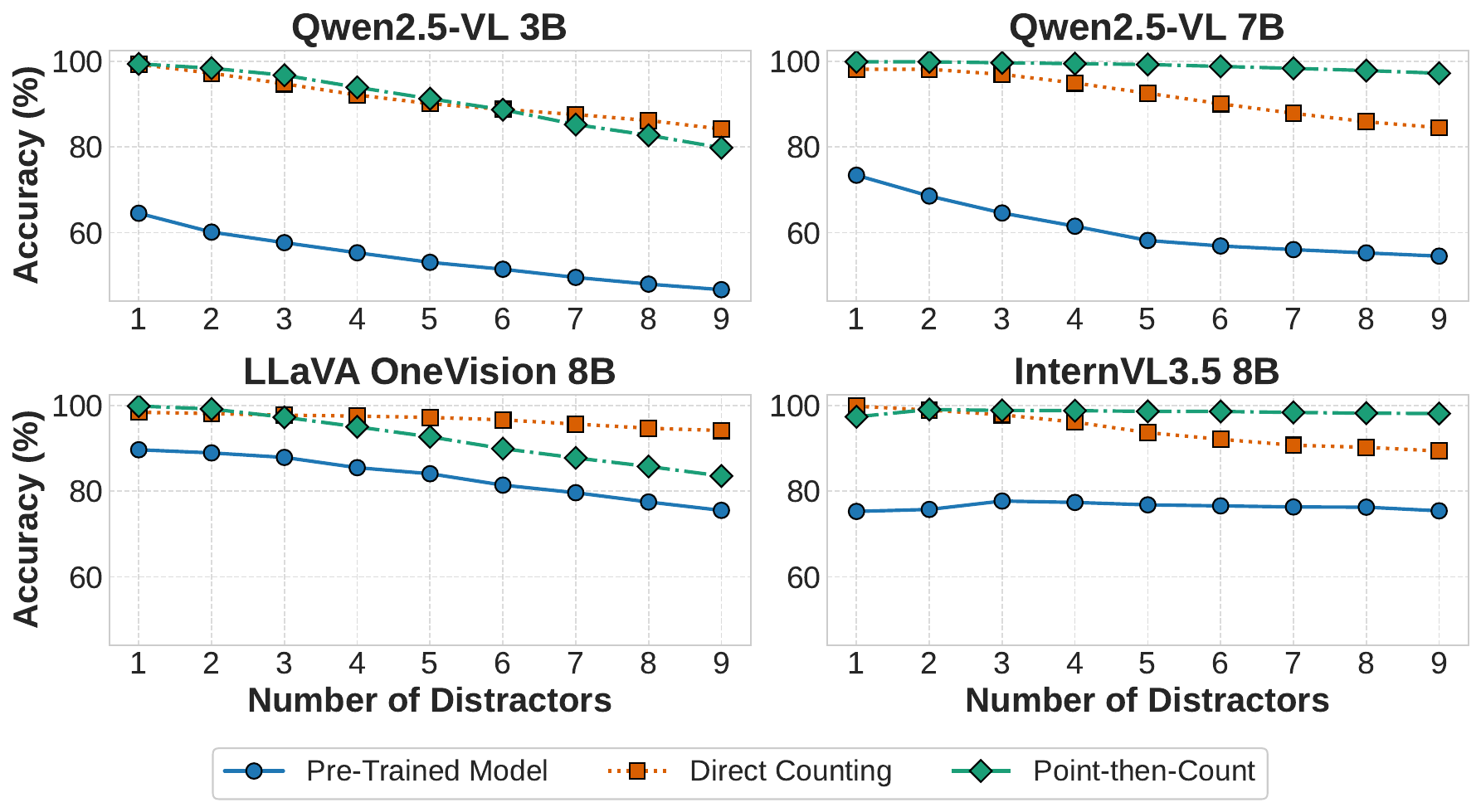}
    \caption{Accuracy (\%) as a function of the number of distractors after fine-tuning on \texttt{Noisy\textsubscript{TR}} for DC or PtC. 
    PtC maintains near-perfect accuracy for \interns and \qwens 7B, while for \qwens 3B and \llavaos PtC degrades faster than DC as distractors increase.
    }
    \label{fig:distractor-acc}
\end{figure*}

\section{Additional Results \& Analyses}
We complement the main paper by providing additional results and analyses.

\begin{table}[t]
    \centering
    \small
    \begin{adjustbox}{width=\columnwidth}
        \begin{tabular}{lcccc}
            \toprule
            & \textit{Circles} & \textit{Squares} & \textit{Triangles} & \textit{Stars} \\
            \midrule
            \textit{R}
            & 98.87 {\scriptsize $\pm$ 1.06}
            & 96.85 {\scriptsize $\pm$ 4.10}
            & 96.06 {\scriptsize $\pm$ 7.61}
            & 99.31 {\scriptsize $\pm$ 0.60} \\
            
            \textit{G}
            & 99.01 {\scriptsize $\pm$ 1.18}
            & 95.58 {\scriptsize $\pm$ 5.68}
            & 96.71 {\scriptsize $\pm$ 5.88}
            & 98.94 {\scriptsize $\pm$ 1.61} \\
            
            \textit{B}
            & \textbf{99.69 {\scriptsize $\pm$ 0.62}}
            & 96.26 {\scriptsize $\pm$ 4.20}
            & 97.46 {\scriptsize $\pm$ 4.04}
            & 99.66 {\scriptsize $\pm$ 0.24} \\
            
            \textit{C}
            & 99.38 {\scriptsize $\pm$ 0.69}
            & 95.71 {\scriptsize $\pm$ 6.71}
            & 96.40 {\scriptsize $\pm$ 6.66}
            & 99.25 {\scriptsize $\pm$ 1.25} \\
            
            \textit{M}
            & 99.14 {\scriptsize $\pm$ 0.76}
            & 93.90 {\scriptsize $\pm$ 8.12}
            & 95.03 {\scriptsize $\pm$ 7.47}
            & 99.35 {\scriptsize $\pm$ 0.49} \\
            
            \textit{Y}
            & 99.01 {\scriptsize $\pm$ 0.70}
            & 96.50 {\scriptsize $\pm$ 4.80}
            & 97.67 {\scriptsize $\pm$ 4.39}
            & 98.80 {\scriptsize $\pm$ 1.86} \\
            \bottomrule
        \end{tabular}
    \end{adjustbox}
    \caption{Accuracy (\%) across object shape and color combinations on the synthetic \texttt{ID} setting for fine-tuned DC models. Values report mean accuracy with standard deviation across models. Results in \textbf{bold} indicate the color-shape combination(s) with the highest accuracy and lower standard deviation. \textit{Notation: \textbf{R}: red, \textbf{G}: green, \textbf{B}: blue, \textbf{C}: cyan, \textbf{M}: magenta, \textbf{Y}: yellow.}}
    \label{tab:shape_color_compositionality-DC}
\end{table}

\begin{table}[t]
    \centering
    \small
    \begin{adjustbox}{width=\columnwidth}
        \begin{tabular}{lcccc}
            \toprule
            & \textit{Circles} & \textit{Squares} & \textit{Triangles} & \textit{Stars} \\
            \midrule
            \textit{R}
            & 99.37 {\scriptsize $\pm$ 0.88}
            & 98.34 {\scriptsize $\pm$ 3.13}
            & 98.01 {\scriptsize $\pm$ 5.41}
            & 99.62 {\scriptsize $\pm$ 0.52} \\
            
            \textit{G}
            & 99.45 {\scriptsize $\pm$ 0.91}
            & 97.70 {\scriptsize $\pm$ 4.36}
            & 98.32 {\scriptsize $\pm$ 4.22}
            & 99.47 {\scriptsize $\pm$ 1.19} \\
            
            \textit{B}
            & 99.78 {\scriptsize $\pm$ 0.42}
            & 97.93 {\scriptsize $\pm$ 3.29}
            & 98.70 {\scriptsize $\pm$ 2.95}
            & \textbf{99.79 {\scriptsize $\pm$ 0.22}} \\
            
            \textit{C}
            & 99.64 {\scriptsize $\pm$ 0.53}
            & 97.79 {\scriptsize $\pm$ 4.92}
            & 98.18 {\scriptsize $\pm$ 4.76}
            & 99.61 {\scriptsize $\pm$ 0.91} \\
            
            \textit{M}
            & 99.55 {\scriptsize $\pm$ 0.67}
            & 96.78 {\scriptsize $\pm$ 6.14}
            & 97.50 {\scriptsize $\pm$ 5.55}
            & 99.66 {\scriptsize $\pm$ 0.46} \\
            
            \textit{Y}
            & 99.45 {\scriptsize $\pm$ 0.66}
            & 98.15 {\scriptsize $\pm$ 3.61}
            & 98.83 {\scriptsize $\pm$ 3.13}
            & 99.38 {\scriptsize $\pm$ 1.37} \\
            \bottomrule
        \end{tabular}
    \end{adjustbox}
    \caption{Accuracy (\%) across object shape and color combinations on the synthetic \texttt{ID} setting averaged across fine-tuned DC and fine-tuned PtC models. Values report mean accuracy with standard deviation across models. Results in \textbf{bold} indicate the color-shape combination(s) with the highest accuracy and lower standard deviation. \textit{Notation: \textbf{R}: red, \textbf{G}: green, \textbf{B}: blue, \textbf{C}: cyan, \textbf{M}: magenta, \textbf{Y}: yellow.}}
    \label{tab:shape_color_compositionality-avg}
\end{table}

\subsection{Compositional Generalization}
\label{subsec:compositional-appendix}
We evaluate compositional generalization across object attributes by training the model on all individual shapes and colors while holding out specific shape-color combinations. At test time, the model is evaluated on these unseen combinations. \cref{tab:shape_color_compositionality-DC,tab:shape_color_compositionality-PtC} report the average accuracy with the standard deviation across models in the \texttt{ID} setting for DC and PtC, respectively. Overall, PtC yields stronger compositional generalization, with two unseen shape-color combinations reaching 100\% accuracy. PtC also exhibits lower variability across object types, with \textit{blue squares} being the most challenging case while still achieving 99.59\% accuracy. By contrast, models fine-tuned with DC reach at most 99.69\% accuracy, down to 93.90\% (for \textit{magenta squares}), and are more sensitive to variations in shape and color. For instance, performance on \textit{Squares} differs by roughly 3 percentage points between \textit{red} and \textit{magenta}. DC also exhibits greater variability across models, with \textit{magenta squares} showing the largest standard deviation at 8.12 percentage points. In contrast, PtC exhibits at most a variability of 0.4 percentage points, suggesting better transferability across models. Finally, we show in~\cref{tab:shape_color_compositionality-avg} that \textit{blue stars} is the color-shape combination with the highest accuracy and lowest standard deviation across models and training approaches. This motivates our choice of \textit{blue stars} as the target object for the \texttt{Noisy\textsubscript{TS}} setting (\cref{subsec:synth-construction-appendix}).

Overall, our findings suggest that PtC promotes stronger compositional generalization across object types than DC.

\begin{table*}[t!]
    \centering
    \small
        \begin{tabular}{lccc ccc ccc}
            \toprule
            \multirow{2}{*}{\textbf{Model}} &
            \multicolumn{3}{c}{\texttt{ID}} &
            \multicolumn{3}{c}{\texttt{OOD}} &
            \multicolumn{3}{c}{\texttt{Noisy\textsubscript{TS}}} \\
            \cmidrule(lr){2-4}\cmidrule(lr){5-7}\cmidrule(lr){8-10}
            & \textbf{F1} & \textbf{EM} & \textbf{Cons.}
            & \textbf{F1} & \textbf{EM} & \textbf{Cons.}
            & \textbf{F1} & \textbf{EM} & \textbf{Cons.} \\
            \midrule
            \textit{\qwens 3B}      &  99.91 &  99.59 & 100.00  & 94.52 & 76.07 & 20.56 & 93.59 & 69.76 & 99.56 \\
            \textit{\qwens 7B}      &  99.89 &  99.55 & 100.00  & 98.59 & 88.05 & 48.79 & 99.19 & 95.77 & 99.75 \\
            \textit{\llavaos 8B} &  99.52 &  98.31 &  99.98  & 96.91 & 80.88 & 77.72 & 89.59 & 70.57 & 99.82 \\
            \textit{\interns 8B}     & 100.00 & 100.00 & 100.00  & 99.89 & 98.06 & 99.27 & 99.84 & 99.05 & 98.51 \\
            \bottomrule
        \end{tabular}
    \setlength{\tabcolsep}{5pt}
    \caption{Coordinate reliability (\%) across three synthetic settings, reported using F1-score, exact match (EM), and consistency (Cons.). Results for \texttt{Noisy\textsubscript{TS}} are computed only on images containing 9 distractors.
    }
    \label{tab:grounding-quality}
\end{table*}

\begin{figure*}[ht]
    \centering
    \includegraphics[width=0.95\linewidth]{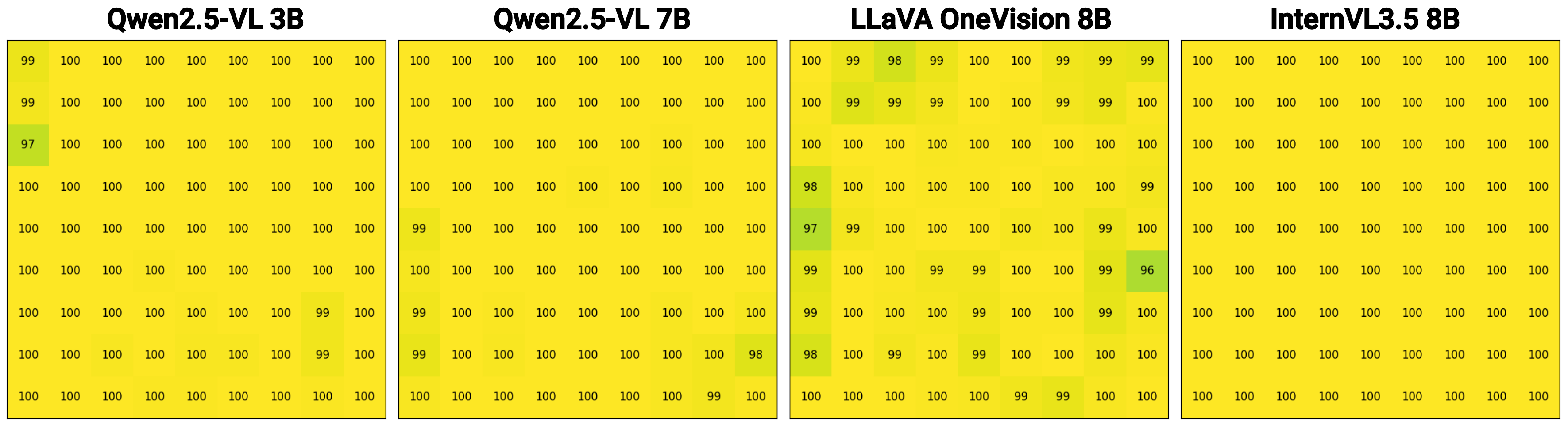}
    \caption{Cell-level F1-score (\%) for each model across our $9\times9$ grid on the synthetic \texttt{ID} setting.}
    \label{fig:cell-level-f1-id}
\end{figure*}

\begin{figure*}[!t]
    \centering
    \includegraphics[width=0.95\linewidth]{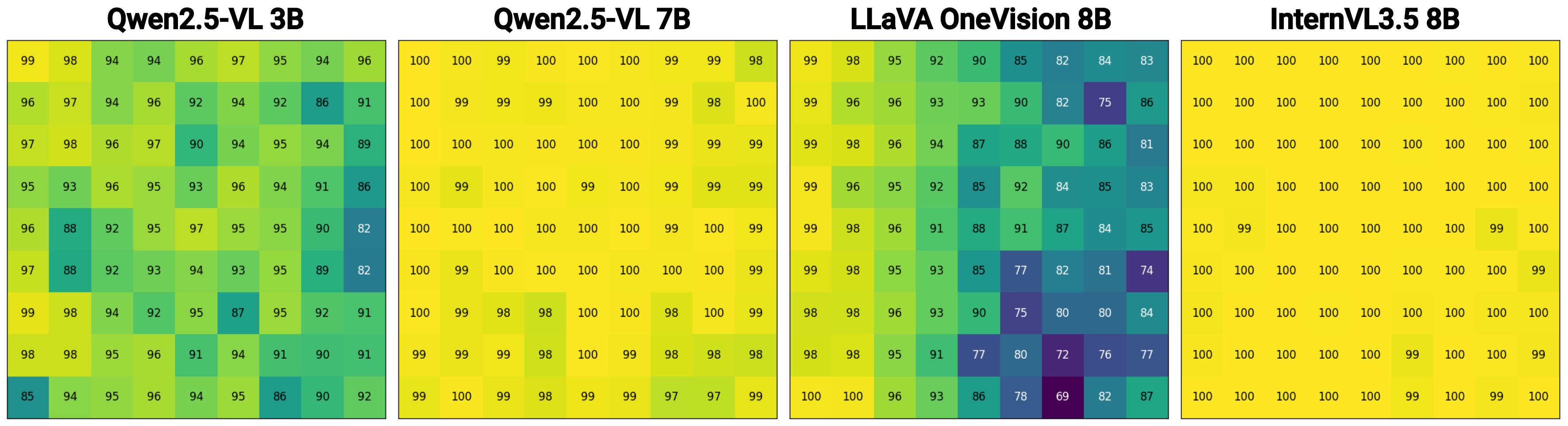}
    \caption{Cell-level F1-score (\%) for each model across our $9\times9$ grid on the \texttt{Noisy\textsubscript{TS}} setting.
    }
    \label{fig:cell-level-f1-noisy}
\end{figure*}

\subsection{Does Point Supervision Improve Robustness to Distractors?}
\label{subsec:distractors-appendix}
To understand which approach is more robust to distractors, we fine-tune the models on a modified set that includes images with up to three distractors (i.e., \texttt{Noisy\textsubscript{TR}}). 
We then evaluate their robustness on images containing up to nine distractors (i.e., \texttt{Noisy\textsubscript{TS}}) and report the results in~\Cref{fig:distractor-acc}.
Our plots show that pre-trained models become less accurate as the number of distractors increases, suggesting that they miscount in the presence of distractors.
The only exception is \intern, which seems to be equally affected by 1 or 9 distactors.
Both DC and PtC improve over the pre-trained models, but point supervision is particularly effective for \interns and \qwens 7B, which maintain more stable results and achieve near-perfect accuracy even with nine distractors.
For the remaining models, point supervision helps up to a moderate number of distractors, but degrades more quickly than DC once distractors become more numerous.

Overall, these results suggest that point supervision can yield more stable performance. However, generalization to a higher number of distractors depends on model architecture and/or (pre-)training data.

\subsection{Grounding}
\label{subsec:grounding-appendix}
We complement the main paper by reporting cell-level F1-scores for the remaining settings. We provide the F1-score, EM, and Consistency for all synthetic settings in~\cref{tab:grounding-quality}. Results show that coordinates are especially reliable in the \texttt{ID} setting, since it matches the training distribution. Notably, most models achieve high and spatially consistent F1-scores across the entire image, as illustrated in~\Cref{fig:cell-level-f1-id}. The only exception is \llavao, which shows slightly reduced grounding performance near the image boundaries.

On \texttt{Noisy\textsubscript{TS}}\footnote{We report results only for images with nine distractors, as this corresponds to the most challenging segment (see~\cref{fig:distractor-acc}).} results show that, despite each model being highly consistent, EM drops substantially for \qwens 3B and \llavao, indicating that most failures stem from localization errors. Cell-level accuracy in~\Cref{fig:cell-level-f1-noisy} reveals that \llavaos exhibits the most pronounced spatial biases in this setting, with performance progressively decreasing from left to right (similar to~\cref{fig:cell-level-f1-ood}). \qwens 3B shows a different trend and achieves a lower F1-score at the bottom and right edges. \qwens 7B exhibits a similar, yet less concerning, pattern, suggesting the possibility of a shared bias within this model family.

\begin{table}[t]
    \centering
    \setlength{\tabcolsep}{8pt}
    \small
    \begin{tabular}{lccc}
        \toprule
        \textbf{Model} & \texttt{ID} & \texttt{OOD} & \texttt{Noisy\textsubscript{TS}} \\
        \midrule
        \textit{\qwens 3B}      & 0.00 & 0.00 & 0.00 \\
        \textit{\qwens 7B}      & 0.00 & 0.09 & 0.03 \\
        \textit{\llavaos 8B}    & 0.01 & 0.08 & 0.20 \\
        \textit{\interns 8B}    & 0.00 & 0.00 & 0.00 \\
        \bottomrule
    \end{tabular}
    \caption{Percentage (\%) of out-of-bounds coordinates predicted by each model across the synthetic \texttt{ID}, \texttt{OOD}, and \texttt{Noisy\textsubscript{TS}} settings. Results for \texttt{Noisy\textsubscript{TS}} are computed only on images containing 9 distractors.}
    \label{tab:out-of-bounds}
\end{table}

Since some models may occasionally produce coordinates that fall outside the image boundaries, which would lower the F1-score for specific spatial regions, we report the percentage of out-of-bounds predictions for each setting in~\cref{tab:out-of-bounds}. Overall, such errors are extremely rare, occurring in fewer than 0.21\% of the cases across all models and settings. In particular, \qwens 3B and \interns never produce out-of-bounds coordinates in any setting, while the remaining models only exhibit negligible rates. Interestingly, these errors appear almost exclusively in the more challenging \texttt{OOD} and \texttt{Noisy\textsubscript{TS}} settings, suggesting that distribution shifts and the presence of distractors slightly increase the likelihood of invalid coordinate predictions.

Overall, these results indicate that invalid coordinate predictions are not a significant source of error in our evaluation, validating the grounding trends reported in~\Cref{fig:cell-level-f1-ood,,fig:cell-level-f1-id,,fig:cell-level-f1-noisy} and confirming that the variations in F1-score primarily reflect the presence of spatial biases.

\subsection{Performance of X-FT on the \texttt{ID} setting}
\label{subsec:xft-appendix}
To show that models fine-tuned to output ``X'' tokens instead of coordinates can still count accurately on in-distribution examples, yet fail to generalize to object counts beyond those observed during fine-tuning (as shown in~\cref{tab:ablations-a}), we evaluate them on the \texttt{ID} setting. The results in~\cref{tab:x-ft-id} show that, although models fine-tuned to output ``X'' instead of coordinates exhibit slightly lower performance than their PtC counterparts ($<3\%$), all models still achieve more than 97\% accuracy. This suggests that, in the \texttt{ID} setting, models can rely on coarse visual cues or on the textual pattern learned during fine-tuning to solve the task even without explicit spatial grounding. However, their accuracy drops substantially on the \texttt{OOD} setting, as shown in~\cref{tab:ablations-a}. Taken together, these results indicate that replacing spatial information with a generic token preserves in-distribution performance, but degrades performance on higher object counts.

\begin{table}[t]
    \centering
    \setlength{\tabcolsep}{10pt}
    \small
    \begin{tabular}{lc}
        \toprule
        \textbf{Model} & \textbf{X-FT} \\
        \midrule
        \textit{\qwens 3B}
        & 98.11 {\scriptsize (\textcolor[HTML]{B22222}{\(\downarrow\) 1.83})} \\
        
        \textit{\qwens 7B}
        & 97.37 {\scriptsize (\textcolor[HTML]{B22222}{\(\downarrow\) 2.51})} \\
        
        \textit{\llavaos 8B}
        & 99.71 {\scriptsize (\textcolor[HTML]{B22222}{\(\downarrow\) 0.05})} \\
        
        \textit{\interns 8B}
        & 99.87 {\scriptsize (\textcolor[HTML]{B22222}{\(\downarrow\) 0.13})} \\
        \bottomrule
    \end{tabular}
    \caption{Accuracy (\%) of each model on the synthetic \texttt{ID} setting after X-FT, where each coordinate is replaced with an ``X'' during PtC training. Values in parentheses indicate the absolute change with respect to the corresponding PtC baseline (\cref{tab:id-synthetic-real-dc-ptc}).}
    \label{tab:x-ft-id}
\end{table}

\begin{figure*}[h]
    \centering
    \includegraphics[width=0.8\linewidth]{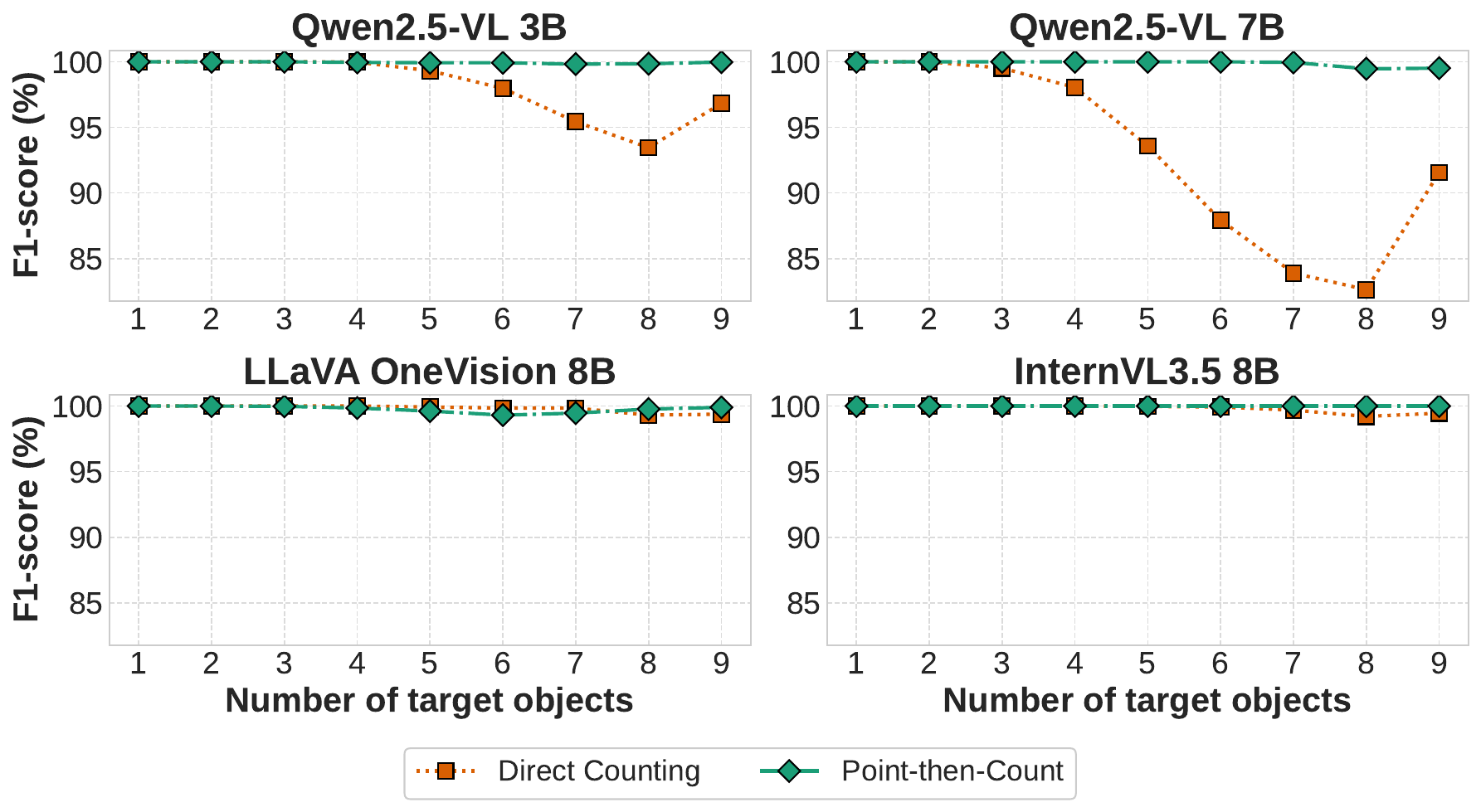}
    \caption{F1-score as a function of the number of target objects under the synthetic \texttt{ID} setting for models fine-tuned with DC or PtC. While \llavaos and \interns show similar performance, \qwens models benefit more from point supervision, suggesting PtC as a more robust choice across object counts.
    }
    \label{fig:baseline-f1}
\end{figure*}

\begin{figure*}[ht]
    \centering
    \includegraphics[width=\linewidth]{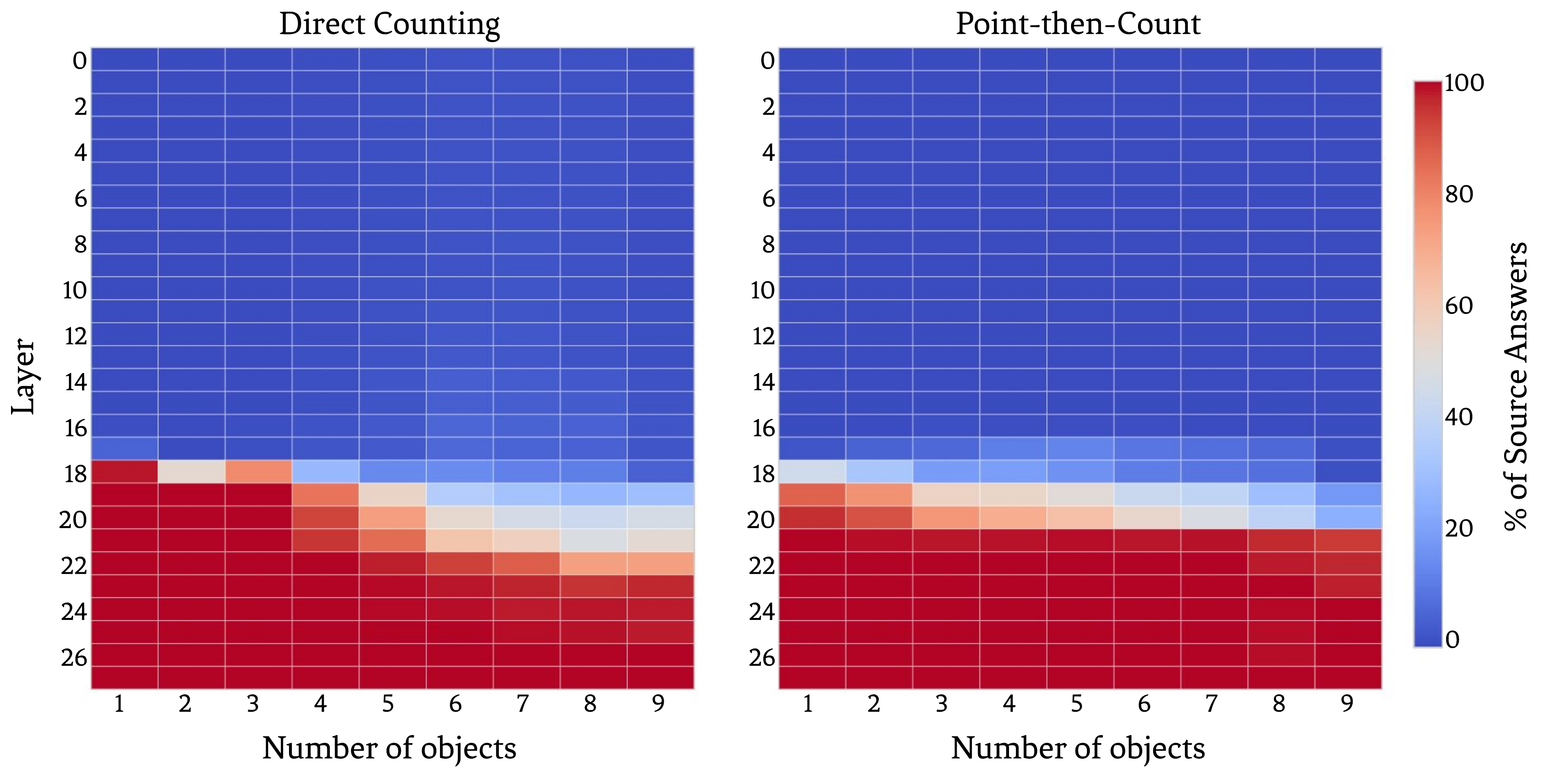}
    \caption{Layer-wise activation patching for \qwens 7B when fine-tuned for DC and PtC. Each cell shows the percentage of source answers after activation patching at layer $L$ (y-axis) for images containing $N$ objects (x-axis). We perform activation patching by substituting the hidden representations of the source image with those of a target image. Low values (blue) indicate that the patched visual representation changes the model's prediction, while high values (red) indicate that the model is unaffected by the patching and outputs the source answer.}
    \label{fig:mechanistic-qwen7b}
\end{figure*}

\begin{figure*}[ht]
    \centering
    \includegraphics[width=\linewidth]{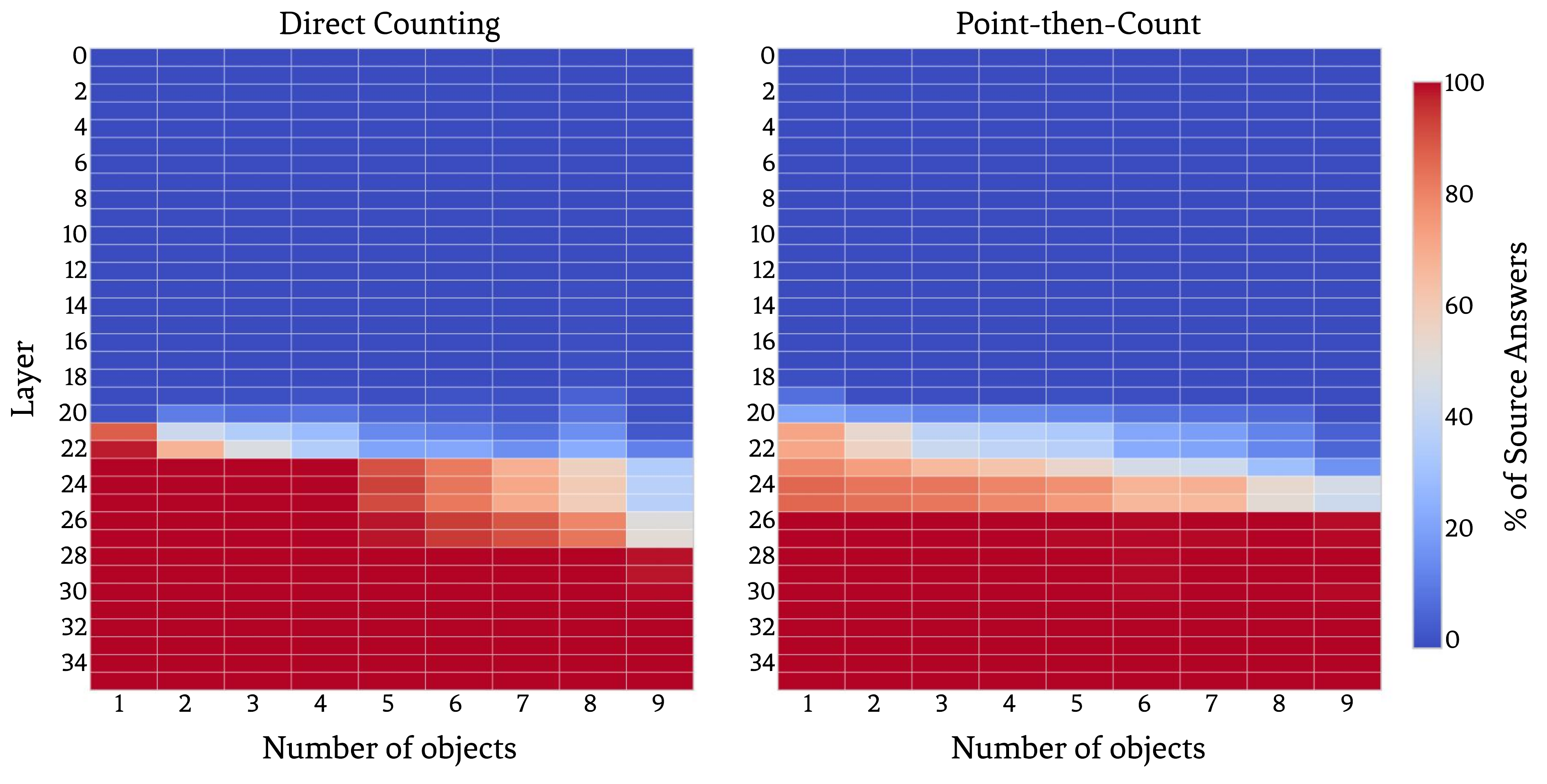}
    \caption{Layer-wise activation patching for \llavaos when fine-tuned for DC and PtC. Each cell shows the percentage of source answers after activation patching at layer $L$ (y-axis) for images containing $N$ objects (x-axis). We perform activation patching by substituting the hidden representations of the source image with those of a target image. Low values (blue) indicate that the patched visual representation changes the model's prediction, while high values (red) indicate that the model is unaffected by the patching and outputs the source answer.}
    \label{fig:mechanistic-llava}
\end{figure*}

\begin{figure*}[ht]
    \centering
    \includegraphics[width=\linewidth]{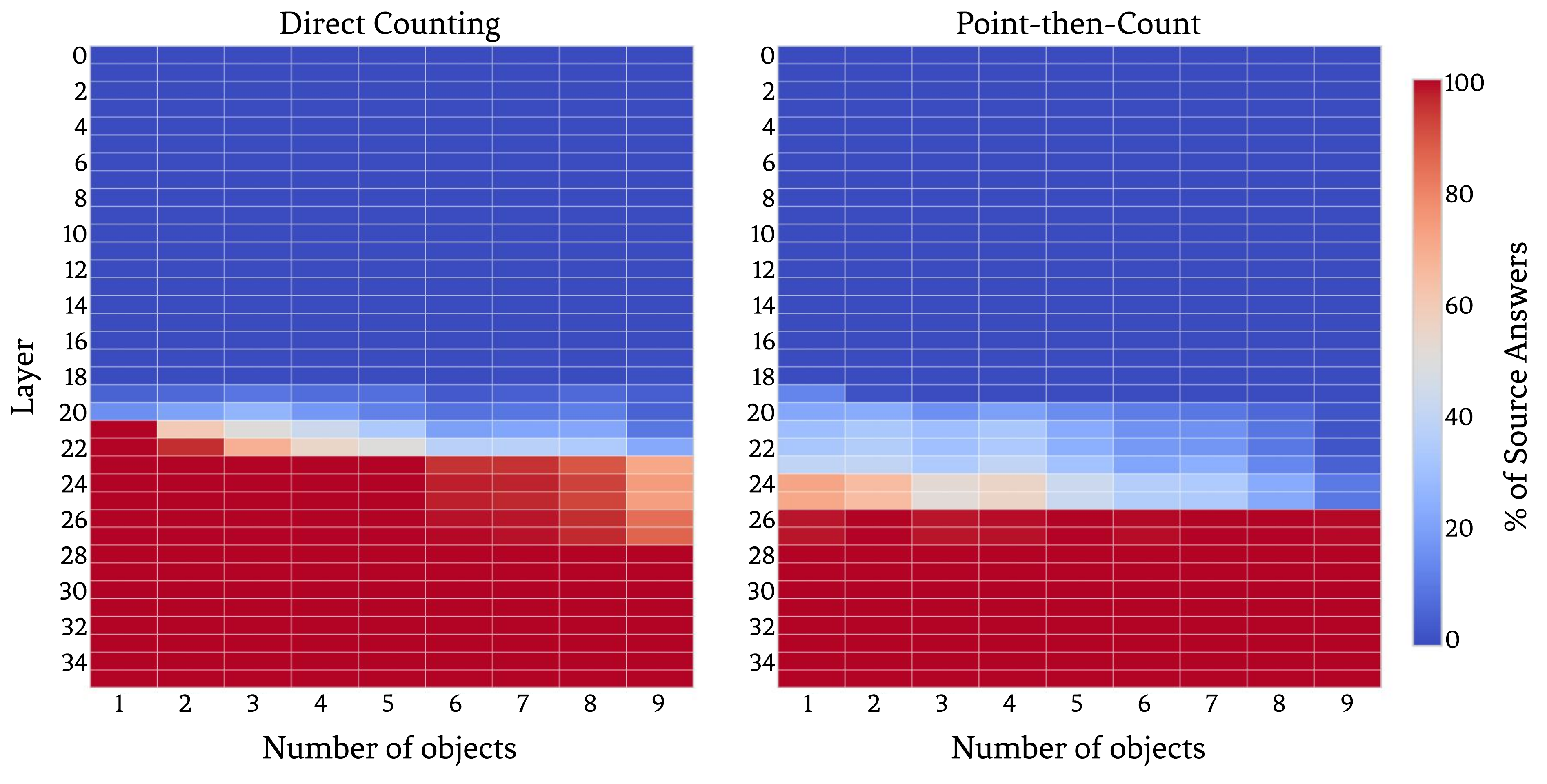}
    \caption{Layer-wise activation patching for \interns when fine-tuned for DC and PtC. Each cell shows the percentage of source answers after activation patching at layer $L$ (y-axis) for images containing $N$ objects (x-axis). We perform activation patching by substituting the hidden representations of the source image with those of a target image. Low values (blue) indicate that the patched visual representation changes the model's prediction, while high values (red) indicate that the model is unaffected by the patching and outputs the source answer.}
    \label{fig:mechanistic-intern}
\end{figure*}

\begin{figure*}[ht]
    \centering
    \includegraphics[width=\linewidth]{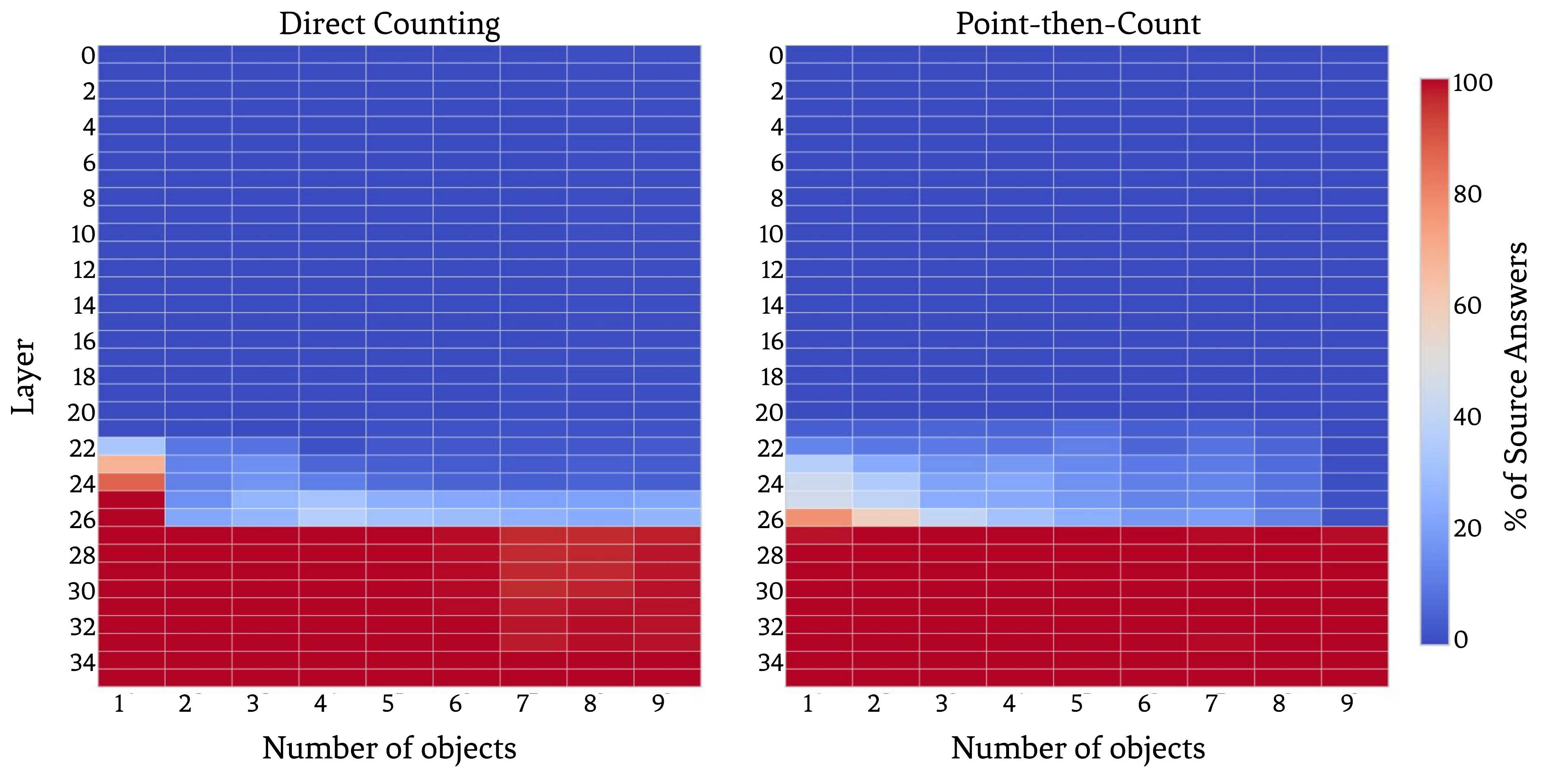}
    \caption{Layer-wise activation patching for \qwens 3B when fine-tuned for DC and PtC. Each cell shows the percentage of source answers after activation patching at layer $L$ (y-axis) for images containing $N$ objects (x-axis). We perform activation patching by substituting the hidden representations of the source image with those of a target image. Low values (blue) indicate that the patched visual representation changes the model's prediction, while high values (red) indicate that the model is unaffected by the patching and outputs the source answer.}
    \label{fig:mechanistic-qwen3b}
\end{figure*}

\subsection{Layer-Wise Activation Patching}
\label{subsec:mechanistic-appendix}
To further analyze when models rely on visual representations during counting, we perform a layer-wise activation patching experiment.
The goal of this analysis is to identify at what layer substituting the visual representation of an input image (i.e., \textit{source}) with that of a related image (i.e., \textit{target}) stops affecting the model's output. Intuitively, if replacing the image representation at a given layer changes the answer towards the target image count, then the model is still using the information from the image. Conversely, if the substitution has no effect, the model is using the information aggregated in the textual tokens.

We construct a set of source-target pairs by stratified sampling from the synthetic \texttt{ID} dataset. Each source image contains \(N\) objects. For each source, we select target images depicting the same object category (e.g., ``blue stars'') but containing a different number of objects. The final set contains a total of 5,184 source-target pairs, computed as 9 count labels, 24 object types, 3 spatial configurations (i.e., source), and 8 related images (i.e., target).

For each source-target pair, we first run the model on the target input and cache the hidden states corresponding to the vision tokens at every transformer layer. We then run the model on the source input. At each layer \(L\), we path the source vision-token hidden states by replacing them with the cached target vision-token hidden states from the same layer, while leaving all other hidden states unchanged. The model then generates an answer from the patched source representation: this allows us to test whether injecting the target visual representation at a given layer changes the model's prediction.

We use this experiment to compare models fine-tuned on DC against models fine-tuned on PtC. Following prior work~\cite{hasani2025understanding}, we report the results for each object count in the range one to nine. \Cref{fig:mechanistic-llava,,fig:mechanistic-intern,,fig:mechanistic-qwen7b,,fig:mechanistic-qwen3b} show, for each layer $L$, the percentage of outputs that match the original source answer. Low values (i.e., blue) indicate that replacing the source vision-token activations with the target changes the model prediction, suggesting that visual information is still influential at layer $L$. Conversely, high values (i.e., red) indicate that the model is unaffected by the patching operation, as it continues to produce the source answer.

Across all models and training approaches, we observe a consistent layer-wise transition. In early and middle layers, the percentage of source-matching outputs is close to zero, indicating that patching the target visual representation affects the model's answer. This suggests that the model is still relying on the image information. In contrast, in the final layers, the percentage of source-matching outputs rises sharply toward 100\%, indicating that replacing the vision-token activations no longer changes the prediction. The exact layer for this transition varies across models and training approaches.

\begin{table}[t]
    \centering
    \small
    \begin{adjustbox}{width=\columnwidth}
        \begin{tabular}{l c ccc}
            \toprule
            \multirow{2}{*}{\textbf{Model}} &
            \textit{TF} &
            \multicolumn{3}{c}{\textit{Fine-Tuning}} \\
            \cmidrule(lr){2-2}\cmidrule(lr){3-5}
            & \textbf{DC}
            & \textbf{DC}
            & \textbf{PtC}
            & \textbf{\# Coord.} \\
            \midrule
            \textit{\qwens 3B}
            & 21.64 & 31.66 & 19.52 & \textbf{81.65} \\
    
            \textit{\qwens 7B}
            & 22.94 & 31.88 & 46.54 & \textbf{94.78} \\
    
            \textit{\llavaos}
            & 19.22 & 31.06 & 72.38 & \textbf{92.11} \\
    
            \textit{\interns}
            & 46.35 & 45.38 & 97.33 & \textbf{98.05} \\
    
            \textit{Gemma 4 E4B}
            & 23.58 & -- & -- & -- \\
            \bottomrule
        \end{tabular}
    \end{adjustbox}
    \caption{Accuracy (\%) on the synthetic \texttt{OOD} setting across training-free (TF) and fine-tuning approaches. \textbf{Bold} values indicate the best result among fine-tuned approaches. Similar to the results in~\cref{tab:synthetic-ood}, computing the count from the coordinates predicted by fine-tuned PtC models yields the highest accuracy.}
    \label{tab:synthetic-ood-appendix}
\end{table}

For \qwens 7B (\cref{fig:mechanistic-qwen7b}), the transition occurs around layers 18 to 22. Under DC, the transition is more gradual and count-dependent, suggesting that counting an increasing number of object may require a larger number of layers, eventually exceeding the model’s computational capacity~\cite{hasani2025understanding}. In contrast, this transition happens approximately at the same layer for PtC, regardless of the object count. A similar pattern is visible for \llavaos (\cref{fig:mechanistic-llava}) and \interns (\cref{fig:mechanistic-intern}), but at different layers. The only exception seems to be \qwens 3B, which requires fewer layers when counting one object, but seems to use the same number of layers for higher counts, even under DC.

Overall, activation patching reveals a common two-stage behavior across models. In the first stage, visual-token activations remain causally important. In the second stage, the answer becomes largely determined by information already integrated into the textual tokens. Regarding the training approach, PtC generally produces a sharper and more count-invariant transition, suggesting that it encourages the model to aggregate visual counting information into a potentially more stable intermediate representation before generating the final count.

\end{document}